\documentclass{article}
 \usepackage[final]{corl_2026} 

\usepackage{enumitem}
\usepackage{booktabs}
\usepackage{multirow}
\usepackage{graphicx}
\usepackage{wrapfig}
\usepackage{amsfonts}
\usepackage{amsmath}
\usepackage{soul}

\title{Learning to Assist: Collaborative VLAs for\\Implicit Human-Robot Collaboration}

\author{
\textbf{Leo Xu}$^{1}$, \textbf{Letian Li}$^{1}$, \textbf{Alex Cuellar}$^{2}$, \textbf{Michael Hagenow}$^{1}$\\
$^{1}$University of Wisconsin--Madison, $^{2}$Massachusetts Institute of Technology\\
\texttt{\{lpxu, lli585, mhagenow\}@wisc.edu}, \texttt{alexcuel@mit.edu}
}

\begin{document}
\maketitle


\begin{abstract}
Human-robot collaboration (HRC) combines the complementary strengths of humans and robots to improve task efficiency. However, many existing collaborative systems rely on hand-engineered pipelines, limiting their scalability and flexibility for new tasks. In this work, we show that models trained end-to-end with imitation learning, specifically vision-language-action (VLA) models, can support collaborative manipulation, and characterize the key factors affecting their real-world performance. We evaluate two state-of-the-art models and identify a failure mode of action-chunking policies in implicit HRC, where demonstration \textit{action leakage} (i.e., action chunks crossing latent task transitions) can cause premature assistive behavior. We find that this issue increases with longer execution horizons and occurs in real-world collaborative VLA systems, such as when a robot attempts to hand over a tool before the person is ready. We propose an inference-time steering method to mitigate these erroneous assistive actions while preserving policy performance. Finally, through a 16-participant user study on a long-horizon collaborative assembly task, we show that steering enables a longer execution horizon while mitigating premature assistance, leading to faster collaboration and fewer failures compared to a shorter-horizon policy. Our paper website can be found at \texttt{\href{https://learning-to-assist.github.io}{learning-to-assist.github.io}}.
\end{abstract}



\section{Introduction}
\label{sec:intro}
Human-robot collaboration (HRC) enables robots to assist humans in tasks where autonomy alone is impractical, but where coordinated and timely robotic assistance can improve efficiency~\cite{ajoudani2018progress}. For example, in a collaborative assembly task, the human may complete subtasks involving advanced reasoning and dexterity, whereas a robot may help by proactively fetching tools and workpieces or completing repetitive motions. To accommodate the complexity introduced by the human, many current HRC systems rely on hand-designed pipelines with targeted modules for intent inference and human-aware motion planning, with learning often used in a limited way~\cite{collab-survey}. While such engineered approaches can be highly effective, they can require significant expertise and effort to design robust behavior that is not easily transferred to new interactions or tasks.

In robot learning, recent large-scale imitation learning approaches have enabled capable autonomous robots across various manipulation skills~\cite{dp,pi0}. Importantly, these methods have shifted robot learning from expert-designed systems toward data-driven systems that can be taught directly through expert demonstrations. A natural question is whether these same approaches can be used to enable more flexible human-robot collaboration. Specifically, we investigate what challenges arise when action-chunking vision-language-action (VLA) policies~\cite{dp,dit-policy,pi0} are trained on collaborative tasks.

\begin{figure}[t]
    \centering
    \includegraphics[width=\textwidth]{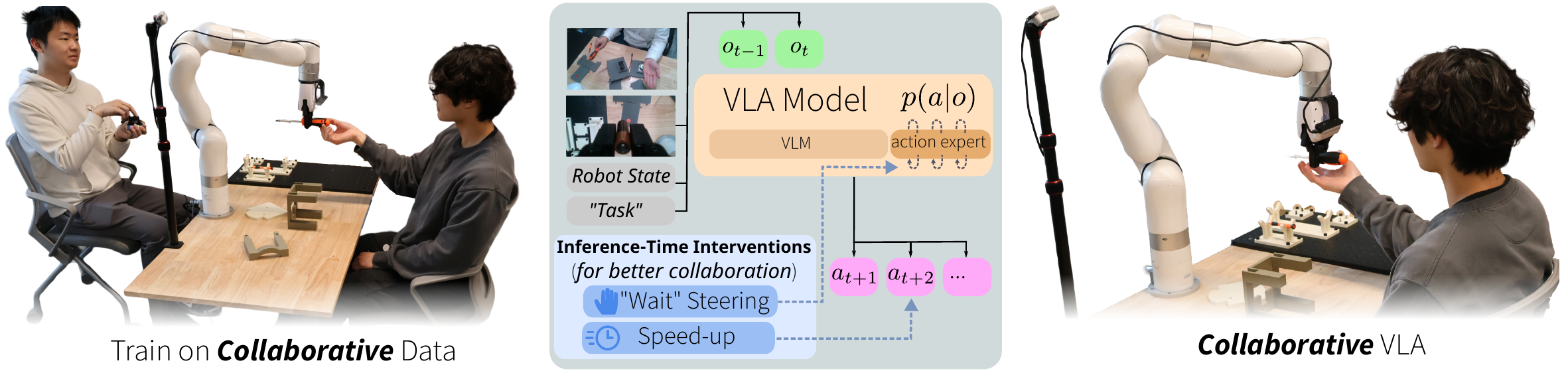}
    \caption{
       Collaborative VLA approach. We augment VLAs with collaborative data and inference-time interventions to improve timing and mitigate premature robot actions in implicit HRC.
    }
    \label{fig:teaser}
    \vspace{-15pt}
\end{figure}

Collaborative settings introduce challenges beyond those in autonomous manipulation. While autonomous approaches primarily emphasize task success and efficiency, collaboration further requires coordinating with the human by attending to signals indicating when and how it should assist. Sometimes these signals are \emph{explicit}, such as when a person asks the robot to grab a workpiece. In highly coordinated teams, this signaling is instead often \emph{implicit}, with the robot anticipating the needs of the human -- e.g., by proactively fetching the appropriate tool depending on where the human is working~\cite{entin1999adaptive,rico2008team}. To enable timely implicit collaboration, a collaborative robot must reason about human intent. However, we find that imitation-learning models trained naively on collaborative demonstrations can produce suboptimal collaborative behavior; they may prematurely commit to actions, causing drift, out-of-distribution states, and failures.

In this paper, we investigate an imitation-learning approach for implicit HRC that leverages VLAs and inference-time steering to mitigate premature assistance while preserving collaborative performance (Figure~\ref{fig:teaser}). We aim to move HRC from hand-designed pipelines to demonstration-driven systems that can be shown how to collaborate effectively with people. Our contributions include:
\begin{enumerate}[leftmargin=*, topsep=2pt, itemsep=2pt, parsep=0pt, partopsep=0pt]
    \item We identify demonstration \emph{action leakage}, a failure mode of action-chunking policies that can cause premature assistive actions (i.e., false starts) in HRC.
    \item We propose an \emph{inference-time steering approach}, leveraging automatically generated basin points, to reduce false starts without retraining or architectural changes.
    \item We assess collaborative VLAs across physical tasks, including model/speed comparisons and a 16-participant user study showing that our steering approach enables longer-horizon execution with reduced time-on-task and fewer failures than a conservative short-horizon baseline.
\end{enumerate}

\section{Related Work}
\label{sec:related-work}

\textbf{End-to-End Imitation Learning. } Imitation learning (IL) enables robots to learn directly from data, such as expert human demonstrations~\cite{argall2009survey}, with recent methods enabling more flexible real-world manipulation~\cite{kawaharazuka2025vision}. While these models can be directly trained on demonstrations (i.e., single-task models), many recent approaches use VLA models~\cite{zitkovich2023rt,kim2024openvla}, which adapt pretrained vision-language models for robot control by adding an action head network and training on a variety of robot manipulation tasks. World action models (WAMs), which often leverage video pretraining and predict future world states, have also gained popularity for strong generalization, but often at greater computational expense~\cite{pai2025mimic,ye2026world}. Broadly, these models enable robots that can perform long-horizon manipulation directly from robot observations (e.g., images and proprioception). We limit our study to VLAs because of their greater open-source availability and lower computational requirements.

Nearly all efforts with VLAs have focused on autonomous capabilities, while effective human interaction is viewed as a core future capability~\cite{bommasani2021opportunities, hagenow2025shared}. While limited past work has explored human-robot interaction with VLAs or similar generative models, the methods typically focus on language interactions~\cite{yell-at-robot, onetwovla, chen2025intentionvla, shi2025hi} or generative co-pilots for assistive teleoperation~\cite{noise-and-back, sun2025flashback, wang2026disco}. Compared to autonomous manipulation settings, HRC with VLAs introduces under-studied challenges related to collaborative timing and robust intent recognition, which we address in our proposed approach.

VLAs commonly output action chunks~\cite{zhao2023learning} containing desired actions for the next $T_p$ timesteps. Prior work has studied action chunking, execution horizon length, and tradeoffs between temporal cohesion and reactivity \cite{dp, mixture-of-horizons, vla-limits}, but does not focus on action leakage that occurs in implicit HRC.

\textbf{Human-Robot Collaboration. }In HRC, a human and robot work together to execute a task~\cite{matheson2019human}. HRC spans many different types of interaction, from collaborative assembly, where the human and robot each perform a set of task steps~\cite{marvel2020towards}, to co-manipulation, which often involves direct physical interaction such as co-carrying or collaborative tool use~\cite{ajoudani2018progress}. Within collaborative assembly, robots can perform subtasks or retrieve workpieces and tools for the human~\cite{ortenzi2021object}. Successful collaboration requires careful attention to timing and the human’s intent~\cite{hoffman2014timing, losey2018review}. While the human can explicitly signal their intent and desired robot assistance, it is often desirable to create proactive robot assistance via implicit signals, such as gaze or the human’s actions, that can enable more effective HRC. Many past methods develop hand-specified pipelines for collaboration, for example, developing explicit modules for intent recognition and human-aware motion planning~\cite{candon2026learning, zhong2026two, cuellar2026multi, fisac2018probabilistically, li2021provably, cuellar2025alignment, huang2015adaptive}.

Learning is often used in a limited capacity in existing methods for HRC~\cite{collab-survey}. While learning methods are used to develop modules in HRC, such as object detection or grasp planning in object handover \cite{kim2024human,duan2022learning}, few past works leverage end-to-end learning for HRC. Diffusion Co-Policy uses a state-conditioned diffusion policy to learn assistance during collaborative table-carrying from human-human demonstrations~\cite{diffusion-copolicy}. The small number of methods leveraging VLAs for HRC, principally for handover interactions, use either a custom input representation tailored to a specific interaction~\cite{li2025robonurse,zhong2026two,robotic-assist} or explicit communication to specify desired robot actions~\cite{wang2026vlabot}. By contrast, we demonstrate how implicit robot assistance can be trained directly from RGB images and further show how this flexible paradigm can enable long-horizon tasks involving mixed collaboration types.

\section{Problem Setting}
\label{sec:problemsetting}
We consider implicit HRC settings where a robot manipulator works with a human to complete collaborative assembly tasks. The robot's objective is to select actions that best support the human. We consider generative robot policies $\pi_\theta$ that produce a distribution over robot actions, $\textbf{a}$, conditioned on robot observations, $\textbf{o}$, a language command, $l$, and the human's latent collaborative intent, $z_h$:
\begin{equation}
\pi_\theta(\textbf{a}|\textbf{o},l,z_h)
\end{equation}
Because the human's true intent is not directly observed, the policy must infer the appropriate assistance from observable context. Equivalently, the robot can be viewed as marginalizing over intent:
\begin{equation}
\pi_\theta(\textbf{a}|\textbf{o},l) = \int \pi_\theta(\textbf{a}|\textbf{o},l,z_h)p(z_h|\textbf{o},l)dz_h
\end{equation}
We assume that the current observation is sufficiently informative to identify the relevant intent. For example, a human extending their hand may indicate the robot should perform a handover, while the human's manipulation of workpieces may signal the next object required. Importantly, these signals can be subtle. We consider action-chunking policies~\cite{zhao2023learning} that output partial trajectories over a prediction horizon, $T_p$, i.e., $\textbf{a}_t\in\mathbb{R}^{n_a \times T_p}$ where $n_a$ is the robot action dimension. We leverage receding-horizon control, where we execute the first $T_a$ (i.e., the execution horizon) samples before replanning~\cite{dp}.
The robot learns its assistive implicit HRC policy from collaborative demonstrations of a given task, $\tau$: $\mathcal{D}_\tau = \left(l_\tau,\{(o_t,a_t)\}_{t=1}^T\right)$, with $T$ total time steps across episodes. We focus on models trained for an individual collaborative task (i.e., all training data has a uniform task label). For implicit HRC, demonstrations include latent intent transitions which do not exist in non-collaborative settings. An observation may still indicate a waiting state, while the corresponding action target spans a subsequently requested assistive motion. We refer to this transition-boundary ambiguity as demonstration action leakage and study its consequences in Section~\ref{sec:false_starts}.

\section{Tasks and Apparatus}

\begin{figure}[t]
    \centering
    \includegraphics[width=\textwidth]{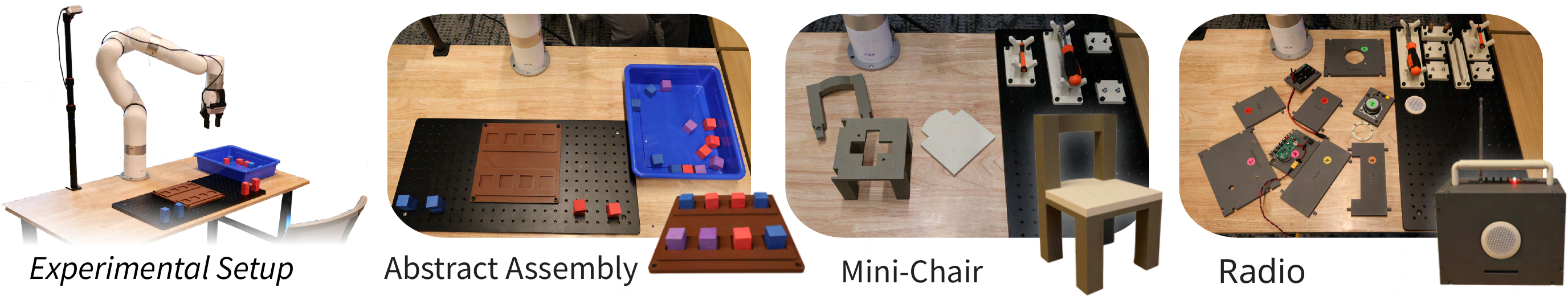}
    \caption{
        Experimental setup and tasks (showing both the \emph{initial state} and the \emph{final assembly}) used in the experiments and the user study.
    }
    \label{fig:tasks}
    \vspace{-15pt}
\end{figure}

We consider three tabletop collaborative assembly tasks designed with increasing complexity, as shown in Figure \ref{fig:tasks}. We investigate two types of robot assistance: sequential steps (the robot must complete a step in response to a human's action) and handovers (the robot handing an object to the human or vice versa).  We refer to our tasks as \textit{Abstract Assembly}, \textit{Mini-Chair}, and \textit{Radio}. 

\textbf{Abstract Assembly} focuses on sequential collaboration with easy-to-manipulate objects, in which the robot and human must each place 4 blocks into slots on a 3D-printed fixture. The human will place either a red, blue, or purple block one by one from left to right. The robot should pick the corresponding color or either color when the person places a purple block (to test if multimodality is preserved). Notably, this first collaborative task does not include physical human-robot interaction.

\textbf{Mini-Chair} incorporates human-to-robot (H2R) and robot-to-human (R2H) handovers. The human attaches the chair seat and back to the chair base, each with two bolts (different types).  The robot hands over bolts and screwdrivers for the task, and must recognize whether the human intends to assemble the seat or back, so it can hand over the appropriate tool and bolts.  After the task, the human hands back both screwdrivers.  This task's handovers and intent recognition (assembling seat vs back) require more flexible behavior and scene comprehension.

\textbf{Radio} is a collaborative assembly task combining sequential and handover steps.  It is longer (an average of 5.5-minute demonstrations) and includes more steps and objects. In total, the robot performs two tool handovers (R2H and H2R), eight object handovers, and two sequential manipulations. This task is used in our user study and is intended to stress test the policies and their ability to perform effectively with human collaborators.

\textbf{Apparatus. } All experiments use the tabletop setup shown in Figure \ref{fig:tasks}. We use a UFactory X-Arm 6 robot with two Intel RealSense D435i cameras: a scene camera and a wrist-mounted camera. The observation includes two RGB images, end-effector (EE) pose, and gripper position. The action space includes EE pose and gripper position, though the specifics differ depending on the model (see Appendix \ref{app:training}). All demonstrations are performed with a seated collaborator and a second person teleoperating with a SpaceMouse. All policy inference is run on an NVIDIA RTX 5090.
\section{Can VLAs Learn Collaborative Tasks?}
\label{sec:rollouts}
Before studying collaboration-specific failure modes, we first establish that VLAs can support collaborative tasks and study the best configuration of model and execution speed for further analysis.

\textbf{Models and Training. } We conduct our experiments using two state-of-the-art models. We trained a Diffusion Transformer model (DiT) based on the Large Behavior Model (LBM) architecture~\cite{dit,dit-policy,lbm}, and fine-tuned the base $\pi_{0.5}$ model~\cite{pi0.5}. By choosing a single-task model and a VLA model pretrained on general manipulation, this experiment also serves to show the impact of general robot pretraining for collaboration. For fairness, both models were trained with a chunk prediction horizon of 16 samples and use an execution horizon of 8 samples (the recommended configuration for LBM). Both models were trained on 100 demonstrations at 10 Hz.  We provide additional training details in Appendix \ref{app:training} and further evaluate $\pi_{0.5}$ under different demonstration budgets in Appendix \ref{app:data}.

\textbf{Speeding Up Robot Execution. } A slow robot may negatively impact productivity by causing the human to wait for the robot to complete its task. Previous work has demonstrated the benefits of faster-than-demonstration robot policy execution~\cite{brown2019extrapolating}. In collaboration, it may not always be desirable to speed up (e.g., a human and robot co-carrying an object), but we consider collaborative actions with sparse physical human-robot interaction where speedup can increase collaborative efficiency. Naively speeding up the robot's execution rate can hurt performance by inducing increased tracking error, distribution shift, and inter-chunk jitter. While previous methods can reduce these issues under faster inference, they require invasive changes at training time~\cite{vlash, sail}. To accommodate these issues, we leverage a simple alignment approach that truncates new action chunks to minimize distance from the last commanded position. The truncated action chunk $\mathbf{a}'_t$ is defined as\footref{indexing}:
\begin{equation}
    \label{eq:pose-alignment}
    \textbf{a}'_t = [a_{t + t_s^*}, ... ,a_{t+T_a-1}], \text{where } t_s^* = \arg\min_{t_s\in \{0..k - 1\}}|| \left[\mathrm{pos}([\textbf{a}_{t}\right]_{t_s})-\mathrm{pos}(a_{t-1})]||
\end{equation} where $k$ defines a maximum truncation length, $\textbf{a}_t$ is the original action chunk, $a_{t-1}$ is the previously commanded action, $\mathrm{pos}(a) \in \mathbb{R}^3$ is the position component of an action, and $\textbf{a}'_t$ is the aligned action chunk. We apply a maximum truncation ($k=3$) to avoid unnecessary truncation when an action chunk crosses a similar pose multiple times. Practically, we find this approach allows us to naively increase the inference rate for more effective robot motions while minimizing jerky robot behavior.

\textbf{Experiments. } We run experiments considering the two models each under a base and speed-up condition, where actions are consumed 1.5x faster. We run 10 policy rollouts for each combination on both \textit{Abstract Assembly} and \textit{Mini-Chair}. All rollouts use a different human collaborator than training. We record metrics related to time, progress (steps completed within the max time of 3 and 6 minutes for \textit{Abstract Assembly} and \textit{Mini-Chair}, respectively), mistakes (recovered by the policy) and failures (requiring a reset).

\textbf{Results. }
The results are shown in Table \ref{tab:policy_performance} with further reporting of a breakdown by specific objects and failure cases in Appendix \ref{app:ext_results}. First, we generally see stronger performance (time and success rate) from $\pi_{0.5}$ compared to the DiT policy. While DiT is mostly successful on the \textit{Abstract Assembly} task, we find that it fails to recognize initial human intent (grabbing a tool once the person starts working on a side of the Mini-Chair) and often fails with the precise fetching manipulations of the tools and bolts. Moreover, DiT struggles to learn the different task orderings in Mini-Chair, where the collaborator can assemble either the back or seat first. Although training data were balanced, DiT succeeded only in one order. Rollouts were evenly split between both orders, yielding a 50\% mean success rate and high SD. These trends suggest that VLA pretraining and recovery data are also beneficial for collaborative tasks, and thus we focus our further evaluation on the $\pi_{0.5}$ model fine-tuned for collaboration. We did not encounter any handover collisions during rollouts.

We also find support for the benefits of model speedup. In \textit{Abstract Assembly}, speedup reduces time considerably with similar rates of mistakes and failures. We see less benefit in \textit{Mini-Chair}, which we attribute to task design. The benefit of speedup is realized over longer robot motions where the human is waiting. In \textit{Abstract Assembly}, the robot must wait until the person selects a block and then must fetch and place an equivalent block. In \textit{Mini-Chair}, however, the robot is often able to proactively fetch a tool or bolt (while the person is working) and only slows the person once they put out their hand for the object, which occurs over a small distance making up a small percentage of the overall task. Because of the limited degradation in accuracy, we argue speedup is often desirable, helping when there are inefficient robot motions that block task progress.
\begin{table}[h]
\centering
\caption{Policy performance across two tasks. Values are reported as mean $\pm$ standard deviation. Bold denotes the best-performing policy for each metric.}
\label{tab:policy_performance}
\scriptsize
\setlength{\tabcolsep}{2pt}
\renewcommand{\arraystretch}{1.08}
\resizebox{\columnwidth}{!}{%
\begin{tabular}{@{}lccccc ccccc@{}}
\toprule
& \multicolumn{5}{c}{\textbf{Abstract Assembly}}
& \multicolumn{5}{c}{\textbf{Mini-Chair}} \\
\cmidrule(lr){2-6}\cmidrule(lr){7-11}
Policy
& \shortstack{Progress $\uparrow$\\(\%)}
& \shortstack{Total $\downarrow$\\time (s)}
& \shortstack{Robot $\downarrow$\\time (s)}
& Mistakes $\downarrow$
& Failures $\downarrow$
& \shortstack{Progress $\uparrow$\\(\%)}
& \shortstack{Total $\downarrow$\\time (s)}
& \shortstack{Robot $\downarrow$\\time (s)}
& Mistakes $\downarrow$
& Failures $\downarrow$ \\
\midrule
DiT
& $95\pm11$ & $120\pm34$ & $109\pm35$ & $0.8\pm0.6$ & $2.1\pm3.7$
& $50\pm53$ & $318\pm56$ & $299\pm68$ & $6.2\pm3.8$ & $0.9\pm1.1$ \\

DiT + Speed
& $\mathbf{100\pm0}$ & $84\pm15$ & $73\pm14$ & $0.7\pm0.7$ & $0.5\pm1.0$
& $50\pm53$ & $275\pm101$ & $264\pm107$ & $7.6\pm5.6$ & $0.3\pm0.5$ \\

$\pi_{0.5}$
& $\mathbf{100\pm0}$ & $71\pm4$ & $58\pm3$ & $\mathbf{0.5\pm0.5}$ & $\mathbf{0.0\pm0.0}$
& $\mathbf{100\pm0}$ & $\mathbf{121\pm6}$ & $\mathbf{101\pm5}$ & $\mathbf{0.1\pm0.3}$ & $\mathbf{0.0\pm0.0}$ \\

$\pi_{0.5}$ + Speed
& $\mathbf{100\pm0}$ & $\mathbf{63\pm3}$ & $\mathbf{51\pm3}$ & $0.8\pm1.1$ & $\mathbf{0.0\pm0.0}$
& $\mathbf{100\pm0}$ & $125\pm9$ & $102\pm9$ & $\mathbf{0.1\pm0.3}$ & $\mathbf{0.0\pm0.0}$ \\
\bottomrule
\end{tabular}%
}
\vspace{-15pt}
\end{table}

\stepcounter{footnote}
\footnotetext{\label{indexing}%
Equation \ref{eq:pose-alignment} is correct for $\pi_{0.5}$, which uses observation $o_t$ consisting of proprioceptive state, task prompt, and images. DiT policy however uses $o_{t-1}, o_t$, producing $\mathbf{a_t} = [a_{t - 1}, ... a_{t+T_a - 2}]$. With alignment, $\mathbf{a}_t' = [a_{t - 1 + t_s^*}, ..., a_{t+T_a - 2}]$. Predicted $\hat{a}_{t - 1}$ is not necessarily previously commanded $a_{t-1}$ due to tracking error.}
\section{Action Leakage in Collaborative VLA Policies}
During qualitative rollout analysis, we observed premature action commitments near interaction boundaries, motivating a more careful analysis of the timing capabilities of VLAs. Concretely, we find that the robot sometimes hands over objects before the person extends their hand. Here we explain the origin of this failure and its mitigation through policy steering. \begin{wrapfigure}{r}{0.36\textwidth}
    \centering
    \includegraphics[width=0.36\textwidth]{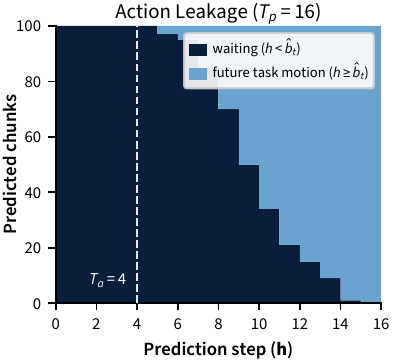}
    \caption{100 Mini-Chair action chunks at a handover boundary sorted by $\hat b_t$. Choosing $T_a \leq b_t$ avoids action leakage.}
    \label{fig:DAL-chunks}
    \vspace{-20pt}
\end{wrapfigure}

\label{sec:false_starts}
\begin{figure}[t]
    \vspace{-30pt}
    \centering
    \includegraphics[width=\textwidth]{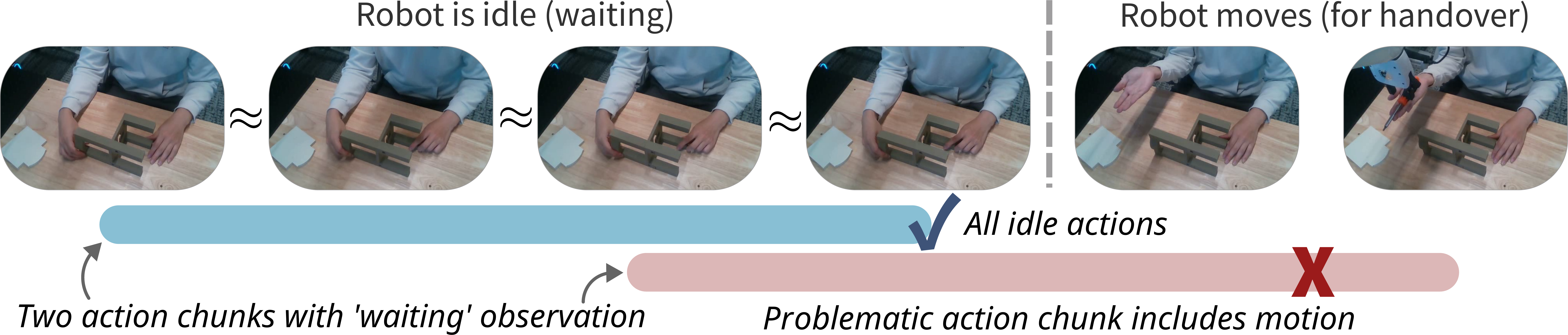}
    \caption{
        Visual depiction of how action chunking from expert demonstrations can lead to \emph{action leakage}, which can cause premature assistive actions in models trained for collaborative tasks.
    }
    \label{fig:false_start}
    \vspace{-13pt}
\end{figure}

\textbf{Premature robot assistance is caused by demonstration action leakage. }
This behavior stems from action-chunking common in VLA models. Within the training data, as illustrated in Figure \ref{fig:false_start}, action chunks may cross transition boundaries (e.g., from waiting to handover) despite observations that the robot should wait. Because states prior to the handover look similar, when the model waits for the human to signal readiness, it may mistakenly generate these problematic action chunks, which include movement toward the workspace, near the boundary of the waiting data. Such outputs produce false starts, inducing drift into out-of-distribution positions and eventually a false commitment (e.g., a premature handover).

We formally define \textit{action leakage} as an action chunk spanning multiple latent human intents. Let $z_t$ denote latent human intent and $\mathbf{a}_t=[a_t,\ldots,a_{t+T_p-1}]$ the target chunk conditioned on $o_t$. Define its latent transition set as $B_t=\{h\in\{1,\ldots,T_p-1\}:z_{t+h}\neq z_{t+h-1}\}$. \emph{Action leakage} occurs when $B_t\neq\emptyset$, i.e., a training chunk spans one or more future intent transitions. At inference, we define $\hat B_t$ analogously as the latent transition set for generated actions, $\hat{\mathbf a}_t$, with the earliest transition $\hat b_t \!=\!\inf \hat{B}_t$. Leakage occurs when $\hat B_t\!\neq\!\emptyset$. Chunks entirely with wrong intent instead have $\hat B_t\!=\!\emptyset$ and constitute normal wrong-mode selection. At a representative handover location in Mini-Chair, we find that all 100 sampled action chunks from $\pi_{0.5}$ with $T_p = 16$ contain these problematic transition boundaries (Figure \ref{fig:DAL-chunks}). Upon removing action leakage chunks from the training set via task-specific heuristics, we find that only $1/100$ trajectories exhibit action leakage and that the policy remains stationary until the collaborator has signaled readiness. This provides direct evidence that action leakage causes false commitments. 

While reducing the model's execution horizon ($T_{a}$) can reduce the prominence of false starts and false commitments, it cannot completely eliminate them (as any action chunk could cross this boundary) and further, a reduced execution horizon can impact temporal consistency (e.g., smoothness) and policy performance \cite{vla-limits, dp, mixture-of-horizons}. Adaptive horizon methods \cite{mixture-of-horizons, vla-limits} also do not inherently solve action leakage as they are unaware of $\hat b_t$ and need additional machinery to inform the proper $T_a$ boundary. Adding additional waiting data to the training set can help reduce false start likelihood, but does not resolve this problem.

\textbf{Mitigating False Starts through Policy Steering. }
We introduce a steering approach to bias the policy sampling toward conservative actions (i.e., waiting), to avoid premature commitment to assistive actions. We find that directly encouraging the model towards waiting behavior (i.e., small velocities) causes issues during slow manipulations. We instead anchor our approach on basin points: waiting positions inferred from demonstrations that we can steer the robot toward during inference. Our approach comprises two false-start safeguarding mechanisms.  We provide additional detail in Appendix \ref{app:steering}.

\underline{FK Steering}. We first augment the policy sampler so that conservative chunks become more likely during inference. To do so, we adapt Feynman-Kac (FK) steering \cite{fk-steering} for VLAs by sampling from the tilted distribution $\pi_{\mathrm{tilt}}(\textbf{a}_t \mid c) \propto \pi_\theta(\textbf{a}_t \mid c)\exp(\lambda r(\textbf{a}_t))$, where $c$ denotes the observation \begin{wrapfigure}{r}{0.36\textwidth}
    \vspace{-0.5em}
    \centering
    \includegraphics[width=0.34\textwidth]{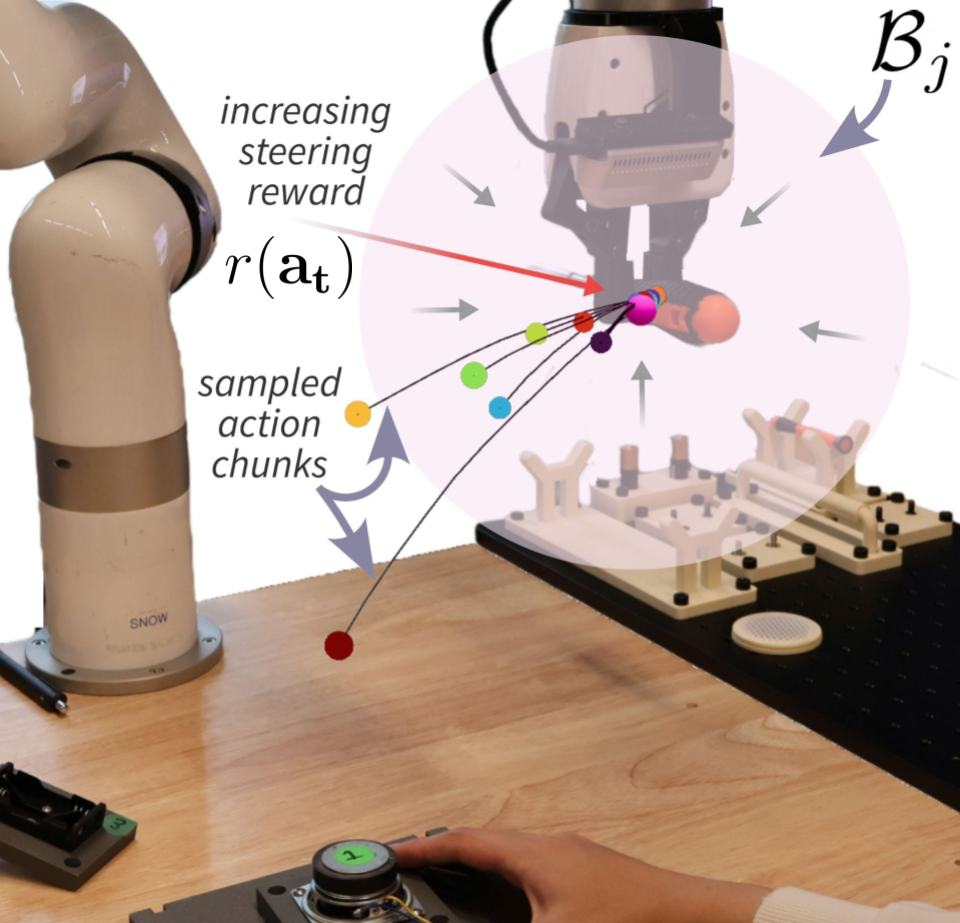}
    \vspace{-5pt}
    \caption{Steering Approach. Steering denoises with higher reward for closer-to-basin trajectories and pulling further encourages waiting.}
    \label{fig:basin_overlay}
    \vspace{-10pt}
\end{wrapfigure}context (camera views, language prompt, and proprioceptive state) and $r(\textbf{a}_t)$ rewards conservative behavior around basin points, $\mathcal{B}_i \in \mathbb{R}^3, i \in\{1, .., N\}$. We approximate the tilted distribution with a sequential Monte Carlo procedure over denoising steps by sampling $M$ candidate action chunks, or particles, ${\{\textbf{a}_t^{(\ell)}} \}_{\ell=1}^{M}$ and assigning particle weights to encourage particles with a high predicted reward $r(\textbf{a}_t)$. Resampling based on these particle weights biases the particle distribution toward conservative chunks while preserving the original base policy diversity. We identify $\mathcal{B}_i$ by applying MeanShift clustering \cite{meanshift} to the training data and define the basin radius, used later to formulate steering reward, to be the 90th percentile cluster radius.

Our reward function $r(\textbf{a}_t)$ consists of three basin-anchored terms. For the current end-effector position $x_{\mathrm{curr}} \in \mathbb{R}^3$, we first identify the active basin as $j=\arg\min_i \|x_{\mathrm{curr}}-\mathcal{B}_i\|_2$. The distance from the active basin is then defined as $d(x)=\|x-\mathcal{B}_j\|_2$. To measure basin membership, we define the soft score $\phi(x)=\sigma(-k(d(x)-m_j))$ when $d(x)\le R_j$, and $\phi(x)=0$ otherwise, where $k$ controls the reward sharpness and $m_j$ is chosen such that $\phi(x)$ approaches zero near the basin boundary $R_j$. Given a sampled action chunk $\textbf{a}_t$, we assign a reward $r(\textbf{a}_{t})=r_{\mathrm{curr}}\cdot r_{\mathrm{traj}}\cdot r_{\mathrm{drift}}$ based on the target end-effector position $x_t$, where $r_{\mathrm{curr}}=\phi(x_{\mathrm{curr}})$, $r_{\mathrm{traj}}=\phi(x_{t+T_p-1})$, and $r_{\mathrm{drift}}=\exp(-s\max(0,d(x_{t+T_p-1})-d(x_{\mathrm{curr}})))$ with $s$ as a hyperparameter. Intuitively, $r_{curr}$ disables steering outside the active basin, $r_{traj}$ favors trajectories that end near the basin, and $r_{\mathrm{drift}}$ penalizes drifting away from it. After the final steering step, we command the highest-reward sample: $\textbf{a}_t^{(\ell^\star)},$ where $ \ell^\star=\arg\max_{\ell\in\{1,\ldots,M\}} r(\textbf{a}_t^{(\ell)}).$ For $\pi_{0.5}$, FK-steering inference takes $88.8 \pm 0.8$ ms with 16 particles versus $67.2 \pm 0.9$ ms normally.

\underline{Basin Pull Steering}. While steering makes sampling conservative actions more likely, the model \begin{wrapfigure}{r}{0.40\textwidth}
    \centering
    \includegraphics[width=0.38\textwidth]{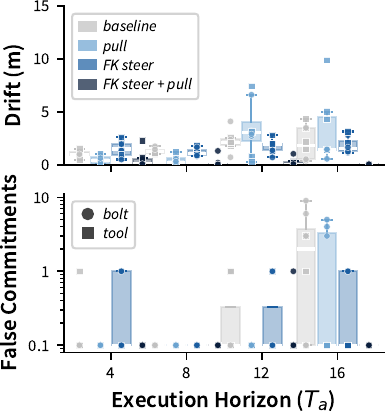}
    \vspace{-5pt}
    \caption{Waiting data experiments. Increased $T_a$ induces false commitments and drift, corrected through steering.}
    \label{fig:fs_plots}
    \vspace{-20pt}
\end{wrapfigure} can still slowly drift outside of $R_j$ due to noise, eventually resulting in false commitments. Therefore, we complement steering with a force that pulls actions toward nearby basins. For the selected action chunk, we interpolate the commanded end-effector position toward the active basin as $\mathrm{pos}(\tilde{a}_{t+h})=(1-w_{\mathrm{pull}})\cdot\mathrm{pos}(a_{t+h})+w_{\mathrm{pull}} \cdot \mathcal{B}_{j}$, for $h=0,\dots,T_p-1$, while leaving the remaining action dimensions unchanged. Here $w_{\mathrm{pull}}=r_{\mathrm{curr}}\left(\max_{\ell} r_{\mathrm{traj}}^{(\ell)}\cdot r_{\mathrm{drift}}^{(\ell)}\right)^p$ and $p\in[0,1]$ is a hyperparameter controlling the strength of the corrective pull. Basin pulling only activates when the robot is within $R_j$ and the sampled particles contain a conservative chunk. Outside $R_j$, the steering reward collapses to zero and the policy samples from the original action distribution. This approach allows the policy to reduce false commitments and drift within $R_j$, while remaining reactive to readiness signals. We also provide a hyperparameter sensitivity analysis in Appendix \ref{app:sensitivity}. 

\textbf{Experiment. }
To study the effects of steering, we run the sped-up $\pi_{0.5}$ for five minutes in a state prior to handover (e.g., holding a screwdriver) with an idle collaborator. We examine false commitments and model drift under different execution horizons with an ablation over the steering interventions. For each condition, we perform eight trials.

\textbf{Results. }
 The results are shown in Figure \ref{fig:fs_plots}. We observe that increasing $T_{a}$ yields more false commitments (e.g., ill-timed handovers) and drift, but basin steering combined with pulling is effective at reducing both issues across execution horizons. Notably, FK steering only activates when the robot is within basin radius, $R_j$. Hence, if the robot drifts outside $R_j$ due to noise, it will tend toward false commitments similarly to the baseline model (as seen with $T_a=4$). For consistency across conditions, our experiments had the collaborator stationary with their hands on the table. We find empirically that false starts occur with greater frequency under user movement, further motivating mitigation. In summary, our approach effectively minimizes errant behavior, while maintaining flexibility to select a task-appropriate $T_a$.
 
\section{User Evaluation}
\label{sec:user_study}
We also conducted a user study to investigate how the collaborative VLA model performs under real user inputs. Our study explores two collaborative VLA configurations informed by the false start mitigation in the previous section. Additionally, the user study allows us to evaluate subjective experience measures of collaborative VLAs, which is critical for future adoption.

\textbf{Participants. } We recruited 16 participants (10M, 6F), aged 19--55 ($M=36.5, SD=12.2$), with minimal-to-moderate robot familiarity (14 \emph{minimal}, 2 \emph{moderate}) from the university campus under a protocol approved by the UW--Madison institutional review board (IRB) 2026-0477.

\textbf{Task and Measures.} Our study focused on the \textit{Radio} task, because it had the highest complexity, longest duration, and included both sequential and handover interactions. We recorded similar metrics to Section \ref{sec:rollouts}, but additionally collected subjective perception through the NASA Task Load Index (TLX)~\cite{hart2006nasa}, System Usability Scale (SUS)~\cite{brooke1996sus}, and relevant sub-scales of the Human-Robot Fluency~\cite{hoffman2019evaluating} questionnaires, which were administered after each condition. We also collected user preferences and conducted a semi-structured interview to elicit user experience feedback.

\textbf{Conditions. } The previous section suggests two mitigations for false-start behavior in collaborative VLAs: leveraging a short execution horizon and applying our steering approach. Given that premature action commitment is highly undesirable in collaboration, we investigate these alternatives as our study conditions to see how the different resulting execution horizons impact collaborative performance and perception. Both conditions use the sped-up fine-tuned $\pi_{0.5}$ policy with $T_p=16$. Condition order was counterbalanced across participants.

\begin{itemize}[leftmargin=5pt, topsep=2pt, itemsep=2pt, parsep=0pt, partopsep=0pt]
    \item \textbf{Shorter Execution Horizon (4)}. Our first condition uses a short execution horizon that minimizes false starts and drift by frequently generating action chunks. The policy is more reactive, but less temporally cohesive. The short execution horizon empirically avoids drift and false starts.
    \item \textbf{Longer Execution Horizon with Steering (12+Steer).} Our second condition uses a longer 12-sample horizon and employs the steering approach in Section \ref{sec:false_starts}. This horizon is less reactive, but more temporally cohesive, which can help with motion legibility and smoothness~\cite{dragan2013legibility}. Our empirical testing indicates that the execution horizon of 12 can induce false starts without intervention.
\end{itemize}

We exclude the 4+Steer and $T_a=12$ conditions since they were either redundant (steering is unnecessary with $T_a=4$) or would not prevent premature commitments ($T_a=12$ without steering). We provide qualitative support for this decision in the following results section.

\begin{figure}[t]
    \centering
    \includegraphics[width=\textwidth]{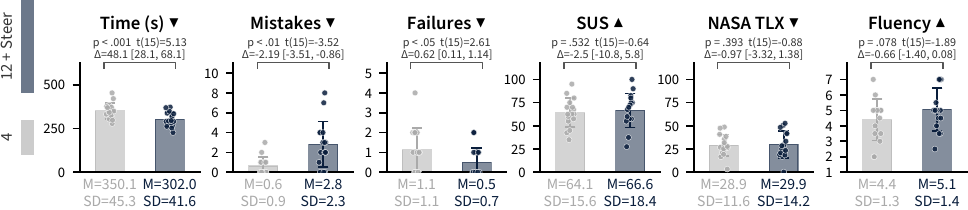}
    \caption{
        User study results.
    }
    \label{fig:user_study}
    \vspace{-15pt}
\end{figure}

\textbf{Results. }
All participants completed the Radio assembly with both conditions. Given our study's exploratory nature, we used uncorrected paired t-tests for our within-subject comparisons and report raw effect sizes with their 95\% confidence intervals. We summarize our quantitative measures in Figure \ref{fig:user_study} and provide additional reporting in Appendix \ref{app:ext_results}. 11 out of 16 participants preferred the 12+Steer condition, mentioning that it was noticeably smoother and faster. The five participants who preferred the shorter execution horizon felt that it was more accurate. 

We found that the 12+Steer condition had significantly more mistakes (e.g., re-grasps) as it is less reactive, but nonetheless achieved significantly shorter time on task and fewer failures requiring reset. Observationally, the shorter horizon took longer to commit to assistance and though the longer-horizon condition had more mistakes, it recovered quickly. During handover, we found that the 12+Steer condition had a significantly higher number of handover collisions (i.e., bumping the participant's hand) ($M=0.5$ vs $M=0.06$, $p<0.05, t(15)=2.2$), though this was not mentioned as a source of discomfort during interviews. We attribute the higher collision rate to the reduced reactivity at $T_a=12$, rather than basin pull distorting handover trajectories as $7/8$ recorded collisions happened outside any basin points (therefore eliminating the effect of basin pull entirely). We did not find significant results in any subjective questionnaires. Overall, these results suggest that our steering approach affords flexibility to select a preferable execution horizon and improve collaborative performance.

To better justify excluding $T_a=12$ without steering, we conducted a small comparison of 12+Steer and $T_a=12$ on the user study task. At a representative handover location, $T_a=12$ suffered from $7 \: (\pm 4)$ premature commitments and a waiting drift of $0.991 \mathrm{m} \: (\pm 0.218 \mathrm{m}, n = 2)$ over five minutes of a stationary collaborator. With steering, we saw $0 \: (\pm 0)$ false commitments and a waiting drift of $0.059 \mathrm{m} \: (\pm 0.002 \mathrm{m}, n = 2)$. We also assessed steering's impact on reaction time: over 50 representative handovers, 12+Steer had a slower average reaction time at $0.403\mathrm{s} \: (\pm 0.259\mathrm{s})$ than $T_a = 12$ with a reaction time of $0.285 \mathrm{s} \: (\pm 0.149\mathrm{s})$. No user study feedback indicated that this overhead impacted collaboration quality.

\section{Discussion \& Limitations}
We investigated VLA models for implicit HRC. Our results suggest that VLAs can be trained to perform collaborative tasks, but implicit collaboration introduces a failure mode in action-chunking VLAs. Demonstration action leakage can cause premature assistance, even when the model is trained on large numbers of demonstrations and waiting data, and inference-time steering offers a practical way to enable longer-horizon, more effective collaborative behavior. Moreover, since steering allows for flexible choices of the reward function, future work could extend basin steering with objectives for personalization or safety.

Our study has limitations that we plan to address in future work. Because we focused on physical HRC, our results are limited to three collaborative assembly tasks (no co-manipulation tasks), and a tractable number of rollouts for in-person experiments on a long-horizon task. In our tasks, intent could be inferred from a single robot observation. Future work should investigate memory approaches for more robust longitudinal intent tracking~\cite{shi2025memoryvla,torne2026mem}. Our primary finding involves premature assistance actions originating from action leakage in VLA action-chunking. In addition to steering, we plan to investigate alternative mitigations (e.g., training-time changes, hierarchical VLAs~\cite{onetwovla}, or autoregressive VLAs~\cite{hu2026ar}). Our study showed that collaborative VLAs still fail and can collide with the human, both of which warrant continued investigation. While implicit collaboration is often desired, we are interested in exploring implicit/explicit mixed approaches as well as how a robot can learn implicit signals from explicit collaboration.


\clearpage
\acknowledgments{We are grateful to Tutor Intelligence for the donation of the robot arm featured in this project. We thank Kerimcan Turkucu for his help with collecting interaction data in each of our tasks and thank Johnny Hong for helping run the demonstration budget Appendix experiments. We also thank William Chen for helpful discussions.}

\bibliography{references}  

\newpage
\appendix
\section{Model Architecture and Training}
\label{app:training}
For all policies, while demonstrations were recorded at 20 Hz, all model training used a downsampled 10 Hz feed and images resized to $224 \times 224$ pixels.
We implement an LBM-style diffusion transformer policy following the Diffusion Policy and LBM action-chunking recipe, adapted for our X-Arm observation and action spaces~\cite{lbm, dit-policy}. As such, we inherit delta pose and delta 6D orientation, observation horizon of 2, prediction horizon of 16, the CLIP ViT visual encoder, hidden dimension 768, 8 transformer blocks, and general structure. All deltas are calculated relative to the first observation in the observation horizon. However, as we train a specialist model for a single task and with a single arm, we omit the language component and any bimanual-specific model inputs and outputs. The training parameters are shown in Table \ref{tab:hyperparams}a.

Our fine-tuned $\pi_{0.5}$ policy uses the standard OpenPi $\pi_{0.5}$ architecture. We cloned the OpenPi GitHub repository (commit 981483d), and fine-tuned the full model from the public $\pi_{0.5}$ base checkpoint. We chose an action space of delta position, absolute quaternion, and absolute gripper position. Concretely, the X-Arm state/action vectors contain $(x, y, z, q_x, q_y, q_z, q_w, g)$ with only Cartesian position dimensions converted to deltas in the action space. These 8D state and action vectors are zero-padded to OpenPi's 32D action dimension. The training parameters are shown in Table \ref{tab:hyperparams}b.

\begin{table}[h]
\centering
\caption{Training hyperparameters.}
\label{tab:hyperparams}
\small
\begin{minipage}[t]{0.48\textwidth}
\centering
\textbf{(a) LBM policy }\\[2pt]
\begin{tabular}{@{}ll@{}}
\toprule
\textbf{Hyperparameter} & \textbf{Value} \\
\midrule
\multicolumn{2}{@{}l}{\textit{Optimization}} \\
Optimizer                    & AdamW \\
Learning rate                & $2\times10^{-5}$ \\
Weight decay                 & $10^{-6}$ \\
$(\beta_1, \beta_2)$         & $(0.9,\ 0.999)$ \\
LR schedule                  & Constant (no warmup) \\
Optimizer steps              & $100{,}000$ \\
Batch size                   & $80$ \\
Grad.\ accumulation steps    & $4$ \\
Effective batch size         & $320$ \\
\midrule
\multicolumn{2}{@{}l}{\textit{Precision \& averaging}} \\
Mixed precision              & bfloat16 \\
EMA power                    & $0.75$ \\
Random seed                  & $42$ \\
\midrule
\multicolumn{2}{@{}l}{\textit{Diffusion}} \\
Training timesteps           & $100$ \\
Beta schedule                & Squared-cosine \\
Prediction target            & $\epsilon$ \\
Inference denoising steps    & $10$ \\
\bottomrule
\end{tabular}
\end{minipage}
\hfill
\begin{minipage}[t]{0.48\textwidth}
\centering
\textbf{(b) \(\boldsymbol{\pi}_{\mathbf{0.5}}\)~policy}\\[2pt]
\begin{tabular}{@{}ll@{}}
\toprule
\textbf{Hyperparameter} & \textbf{Value} \\
\midrule
\multicolumn{2}{@{}l}{\textit{Optimization}} \\
Optimizer              & AdamW (OpenPi default) \\
Gradient clipping      & $1.0$ \\
$(\beta_1, \beta_2)$   & $(0.9,\ 0.95)$ \\
Weight decay           & $10^{-10}$ \\
Adam $\epsilon$        & $10^{-8}$ \\
LR schedule            & Cosine \\
Warmup steps           & $1{,}000$ \\
Peak learning rate     & $2.5\times10^{-5}$ \\
Decay steps            & $30{,}000$ \\
Final learning rate    & $2.5\times10^{-6}$ \\
Optimizer steps        & $25{,}000$ \\
Global batch size      & $32$ \\
\midrule
\multicolumn{2}{@{}l}{\textit{Precision \& averaging}} \\
Mixed precision        & bfloat16 \\
EMA decay              & $0.99$ \\
Random seed            & $42$ \\
Checkpointing          & Final step \\
\bottomrule
\end{tabular}
\end{minipage}
\end{table}

\section{Impact of Collaborative Training Data on Performance}
\label{app:data}
In the experiments in Sections \ref{sec:rollouts}--\ref{sec:user_study}, we trained the collaborative VLA models on 100 demonstrations to give sufficient data for the single-task model. From our results, we conclude that leveraging the general manipulation pretraining of $\pi_{0.5}$ leads to broadly better collaborative performance. However, it remains unclear how much collaborative data is needed to reliably complete the task. Hence, we ran an experiment to look at how performance changes across different numbers of collaborative demonstrations for our most complex task: Radio assembly. Fewer required demonstrations would greatly improve the utility of collaborative VLAs in practice. 

\begin{figure*}[t]
    \centering
    \begin{minipage}[b]{0.48\textwidth}
        \centering
        \includegraphics[width=\textwidth]{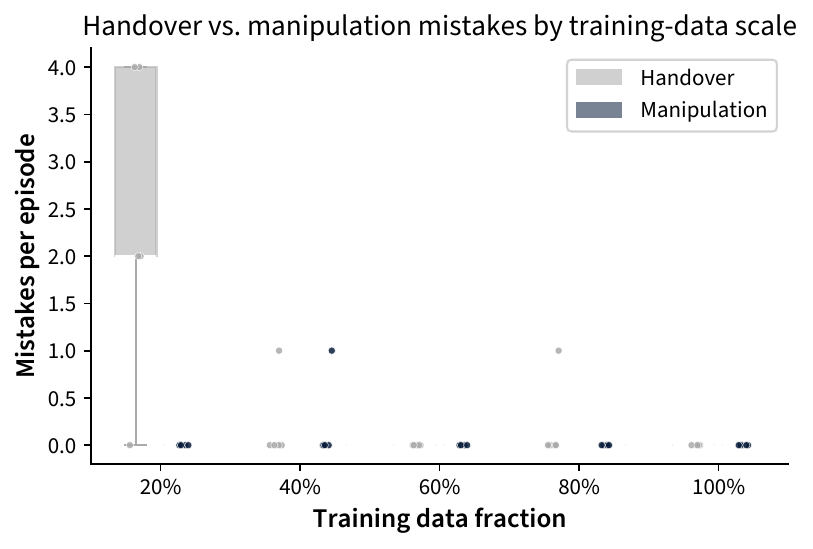}
        \caption{Mistakes by subtask: handover vs.\ manipulation.}
        \label{fig:data-mistake-scale}
    \end{minipage}
    \hfill
    \begin{minipage}[b]{0.48\textwidth}
        \centering
        \includegraphics[width=\textwidth]{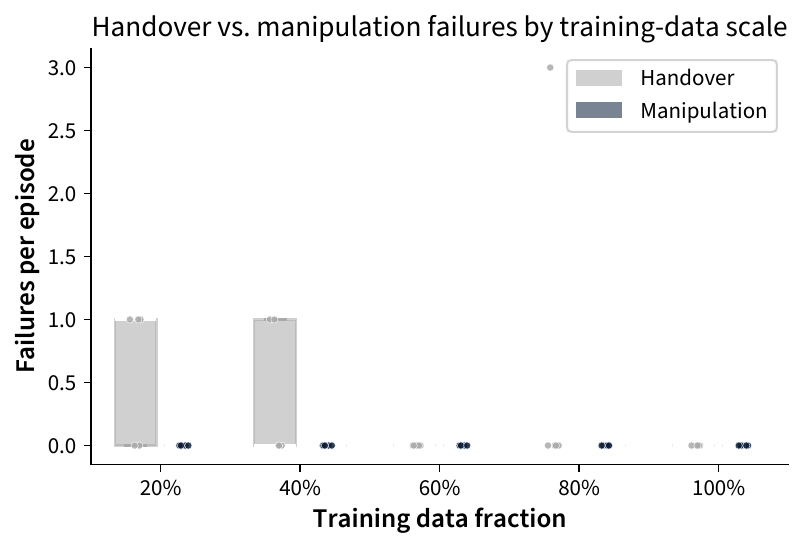}
        \caption{Failures by subtask: handover vs.\ manipulation.}
        \label{fig:data-failure-scale}
    \end{minipage}
    \vspace{1.0em}
    \begin{minipage}[b]{0.48\textwidth}
        \centering
        \includegraphics[width=\textwidth]{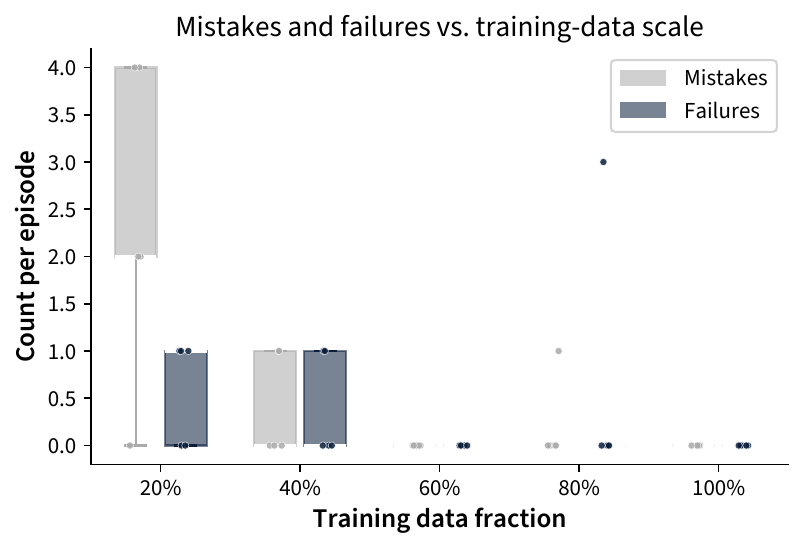}
        \caption{Total mistakes and failures per episode vs.\ training-data fraction.}
        \label{fig:data-scale}
    \end{minipage}
\end{figure*}

\textbf{Experiment. }
We fine-tuned five different $\pi_{0.5}$ models each with increasing amounts of training data in increments of 20 demonstrations and examine the overall trend in observed failures and mistakes in Figure \ref{fig:data-scale}. For each trained model,  we collected five rollouts with a collaborator not in the training set. As seen in Figures \ref{fig:data-mistake-scale} and \ref{fig:data-failure-scale}, the majority of mistakes and failures arose from handovers rather than manipulation inaccuracies. Notably, even with only 20 demonstrations, $\pi_{0.5}$ makes no manipulation errors. We attribute this to $\pi_{0.5}$'s large-scale manipulation pretraining, whereas we hypothesize there is no pretraining on human-robot collaboration (e.g., handovers). We observe that it took up to 60 demonstrations for $\pi_{0.5}$ to learn handover behavior reliably. Hence, we conclude that VLAs like $\pi_{0.5}$ can quickly adapt to collaborative tasks involving mostly manipulation, but require more data to learn the interactive portions of HRC. It may also be possible to pre-train future models on general HRC data, which we hypothesize could greatly reduce task-specific demonstrations.

\section{Extended Discussion of Policy Steering}
\label{app:steering}

\textbf{Action Chunking \& Action Leakage.}
Action chunking was originally popularized by Zhao et al.~\cite{zhao2023learning} where chunking was used to address issues with non-Markovian pausing and temporally correlated noise \cite{swamy2022causal} within the training distribution. Without action chunking, Zhao et al. found that the policy could pause indefinitely in certain states due to similar observations while action chunking allowed the policy to learn to pause for $k$ timesteps before moving at timestep $t + k$, preventing the robot from freezing.

However, we find that action chunking \emph{introduces the opposite problem in HRC} where the policy prematurely moves instead of learning to fully wait. In other words, action chunking enables behavior where the policy can forecast future movement before an exogenous signal for action is seen. We note that this issue is less prevalent in fully autonomous tasks than in HRC as timing based on external signals is comparatively limited. 

\textbf{Feynman-Kac Steering Clarifications.}
We introduce policy steering to bias inference toward conservative action chunks until an external movement signal is observed, using Feynman-Kac (FK)-style resampling \cite{fk-steering} to approximate a reward-tilted particle population corresponding to $\pi_{\mathrm{tilt}}(\mathbf{a}_t \mid c) \propto \pi_{\theta}(\mathbf{a}_t \mid c)\exp(\lambda r(\mathbf{a}_t))$, where $c$ denotes the observation context and $\lambda \in \mathbb{R}$ controls the steering temperature.

First, for $\tau \in \{0, \ldots, S\}$, we construct a potential function
$
G_\tau(\mathbf{a}_t^{(S:\tau)}) = \exp(\lambda (r(\hat{\mathbf{a}}_t^{(\tau)}) - r(\hat{\mathbf{a}}_t^{(\tau + 1)})))
$
where $\tau$ represents each step of denoising or the Euler method for flow-matching models and $\hat{\mathbf{a}}_t^{(\tau)}$ is the estimated denoised sample from an intermediate denoising step $\tau$. For $\pi_{0.5}$, which uses flow matching, $\hat{\mathbf{a}}_{t}^{(\tau)} = \mathbf{a}_t^{(\tau)} + \frac{\tau}{S} u_t^{\theta}(\mathbf{a}_t^{(\tau)} \mid c)$, and for the DiT policy we use the DDIM scheduler to estimate the fully denoised particle. At $\tau = S$, we take $G_S(\mathbf{a}_t^{(S)}) = \exp(\lambda r(\hat{\mathbf{a}}_t^{(S)}))$. We use the convention of choosing $\mathbf{a}_t^{(0)}$ to denote the fully denoised action chunk and $\mathbf{a}_t^{(S)}$ to be the action chunk drawn from $\pi_{\mathrm{prior}} = \mathcal{N}(0, 1)$. This choice of potential function has the property that:

\begin{align}
\prod_{\tau = 0}^{S} G_\tau(\mathbf{a}_t^{(S:\tau)})
&= \exp\left(\lambda \left(r(\hat{\mathbf{a}}_t^{(S)}) + \sum_{\tau = 0}^{S - 1} \left[r(\hat{\mathbf{a}}_t^{(\tau)}) - r(\hat{\mathbf{a}}_t^{(\tau + 1)})\right]\right)\right) \\
&= \exp(\lambda r(\mathbf{a}_t^{(0)}))
\end{align}
which reduces to the terminal tilt. This property allows us to apply the terminal reward tilt through intermediate denoising/integration potentials, so that iterative reweighting targets the desired $\exp(\lambda r(\mathbf{a}_t^{(0)}))$. At step $\tau = 0$, $\hat{\mathbf{a}}^{(0)}_t = \mathbf{a}^{(0)}_t$, since the action chunk has been fully denoised after all steps $\tau > 0$. To estimate the tilted distribution iteratively, we sample $M$ particles $\{\mathbf{a}_{\ell, t}\}_{\ell = 1}^{M}$ and assign each particle a weight at intermediate denoising/integration steps:
$$w_\ell^{(\tau)} = \frac{G_\tau(\mathbf{a}_{\ell, t}^{(S:\tau)})}{\sum_{q=1}^{M}G_\tau(\mathbf{a}_{q, t}^{(S:\tau)})}$$

We then perform multinomial resampling at every step $\tau$, where each particle is redrawn with probability $w_\ell^{(\tau)}$. At step $\tau = 0$ (i.e., after steering), we command the particle with index $\arg \max_{\ell} r(\mathbf{a}_{\ell, t}^{(0)})$. While FK steering can exhibit particle collapse in flow-matching models due to their determinism after initial sampling, resulting in behavior similar to Best-of-N sampling~\cite{kac-flow}, in practice, we find that desired robot assistive behavior is preserved.

\textbf{Properties of Reward Design.} We design our reward function such that it can both mitigate false starts and leave the base policy unchanged outside the active basin. For both $w_{\mathrm{pull}}$ and $r(\mathbf{a}_t)$, we include $r_{\mathrm{curr}}$ by design. Since $r_{\mathrm{curr}}$ is zero when $d(x_{\mathrm{curr}}) > R_j$, $w_{\mathrm{pull}}$ and $r(\mathbf{a}_t)$ are both clamped to zero. This effectively allows the model to behave normally outside of $R_j$ as $\forall \ell \in \{1, ... , M \}$, $\exp(\lambda r(\hat{\mathbf{a}}_{\ell, t}^{(\tau)})) = \exp(\lambda 0) = 1$, inducing uniform resampling across particles at each step $\tau$.

\textbf{Basin Steering Hyperparameters.}
We use the same steering hyperparameters for all experimental tasks (see Table \ref{tab:basin_hyperparams}), showing the relative robustness of the steering method across different tabletop manipulation tasks. However, choices of hyperparameters do fundamentally adjust robot behavior. For instance, choosing an aggressive $k$ can push most rewards to zero if particles are slightly away from the basin $B_j$, leading to no resampling at intermediate steps. Moreover, setting $p$ too low can lead the model to never leave the basin point. Finally, while increasing the number of particles provides a better approximation of the policy distribution, more particles will incur additional runtime overhead. We found that $M = 16$ particles incurred a modest overhead of 20 ms on an RTX 5090 while providing enough of an approximation of the underlying distribution for effective steering. 

\begin{table}[h]
\centering
\caption{Basin steering hyperparameters used in our experiments.}
\label{tab:basin_hyperparams}
\small
\setlength{\tabcolsep}{5pt}
\renewcommand{\arraystretch}{1.12}
\begin{tabular}{@{}p{0.14\linewidth}p{0.16\linewidth}p{0.62\linewidth}@{}}
\toprule
Parameter & Value used & Effect \\
\midrule
$M$ & 16 & Number of candidate action chunks used for FK steering. \\
$\lambda$ & 2 & Strength of the exponential tilt $\exp(\lambda r(\mathbf{a}_t))$. \\
$k$ & 100 & Sharpness of the soft basin-membership score $\phi(x)$. \\
$p$ & 0.05 & Pull strength toward the active basin, with $p\in[0,1]$. \\
$s$ & 2 & Penalty scale for action chunks whose endpoints drift away from the active basin. \\
\bottomrule
\end{tabular}
\vspace{-6pt}
\end{table}

\section{Basin Steering Sensitivity Analysis}
\label{app:sensitivity}

\begin{figure}[h]
    \centering
    \includegraphics[width=\textwidth]{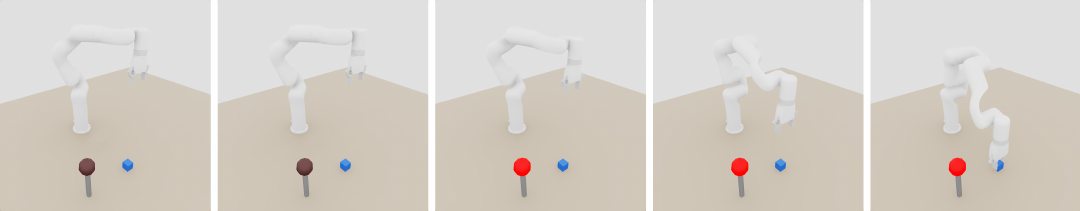}
    \caption{
        A training episode of our Isaac Sim task where the robot must wait before grabbing the block. 
    }
    \label{fig:sim-task}
    
\end{figure}

To tractably assess the sensitivity of basin steering to hyperparameters, we designed a task in Isaac Sim \cite{gao2026nvidia} that elicits action leakage similar to handover boundaries as seen in Figure \ref{fig:sim-task}. In this environment, the robot must wait until a light turns on upon which it should grab the blue block, ending the episode. The time in seconds when the light turns on is given by $T_{\mathrm{on}} \sim \mathrm{Exp}(1/6)$ and the robot reaction time by $T_{\mathrm{react}} \sim \mathrm{Uniform(0.2, 0.5)}$. The robot is initialized in a starting pose $x_{\mathrm{start}}$ with noise added to each Cartesian coordinate distributed by $\mathcal{N}(0, 0.01)$. Once $t \geq T_{\mathrm{on}} + T_{\mathrm{react}}$, we linearly interpolate the end-effector pose $x_{\mathrm{curr}}$ with the block pose $x_{\mathrm{block}}$.

\begin{figure}[h]
    \centering
    \includegraphics[width=\textwidth]{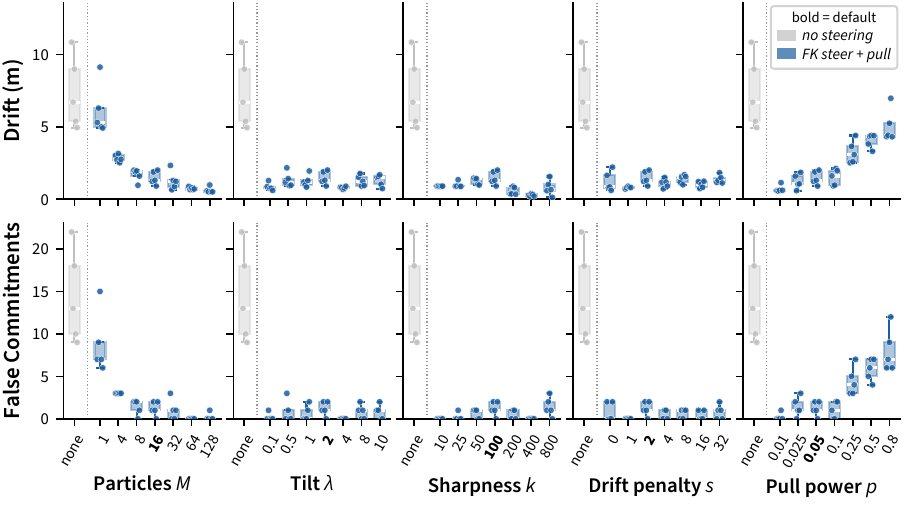}
    \caption{
        ``One-at-a-time" hyperparameter sweep of basin steering compared with no steering.
    }
    \label{fig:sweep}
    
\end{figure}

\begin{figure}[h]
    \centering
    \includegraphics[width=\textwidth]{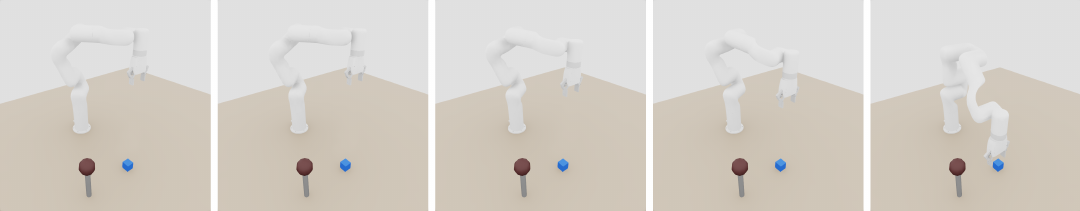}
    \caption{$\pi_{0.5}$ exhibiting action leakage at inference time while the light is off.}
    \label{fig:sim-task-rollout}
    
\end{figure}

We collected 200 episodes and trained $\pi_{0.5}$ on this task. During inference, we never turn the light on and record waiting drift along with the number of false commitments over the span of five minutes. We found that without steering, $\pi_{0.5}$ suffered from action leakage similarly to our real-world findings and show an example rollout in Figure \ref{fig:sim-task-rollout}. Quantitatively, we ran a ``one-at-a-time" hyperparameter sweep where we swept each hyperparameter while keeping all others at their default values from Appendix \ref{app:steering}. Results for each hyperparameter combination were collected over five trials of five minutes and are shown in Figure \ref{fig:sweep}.

We find that basin steering is most sensitive to the number of particles sampled $M$ and the basin pulling power $p$. We caution against setting $p$ too low as this can lead to behavior where the policy never escapes the basin point. If computationally feasible, sampling more particles $M$ at inference time can lead to fewer false commitments. 

\section{Extended Experimental Results}
\label{app:ext_results}

\subsection{Model Rollouts}
In our rollouts, we distinguish between mistakes and failures, where failures consist of erroneous actions requiring human intervention (resetting the robot to a predetermined task-specific position) and mistakes represent inaccuracies during operation that do not require resets. We define handover collisions as moments when the robot touches an outstretched hand during handover. During all rollouts and our user study, handover collisions did not result in any injury to the collaborator. Wrong commitments (e.g., grabbing the wrong screwdriver) are an aggregate measure recording incorrect actions whether recoverable or not. We separate this measure into commitment mistakes and commitment failures accordingly when considering aggregate failures or mistakes.

We show both disaggregated mistake and failure counts for Abstract Assembly and Mini-Chair in Table \ref{tab:policy_performance_abstract_full} and Table \ref{tab:policy_performance_handover_full}, respectively.

\begin{table}[h]
\centering
\caption{Abstract Assembly --- all individual metrics. Values are mean $\pm$ SD. Bold = best per column.}
\label{tab:policy_performance_abstract_full}
\scriptsize
\setlength{\tabcolsep}{2pt}
\renewcommand{\arraystretch}{1.08}
\resizebox{\columnwidth}{!}{%
\begin{tabular}{@{}lcccccccccc@{}}
\toprule
Policy
& \shortstack{Progress $\uparrow$\\(\%)} & \shortstack{Total time $\downarrow$\\(s)} & \shortstack{Robot time $\downarrow$\\(s)} & \shortstack{Human time $\downarrow$\\(s)} & \shortstack{Wrong\\commit. $\downarrow$} & \shortstack{Grasp\\mistake $\downarrow$} & \shortstack{Place\\mistake $\downarrow$} & \shortstack{Grasp\\failure $\downarrow$} & \shortstack{Place\\failure $\downarrow$} & \shortstack{False\\starts $\downarrow$} \\
\midrule
DiT
& $95\pm11$ & $120\pm34$ & $109\pm35$ & $12\pm2$ & $\mathbf{0.0\pm0.0}$ & $\mathbf{0.0\pm0.0}$ & $0.8\pm0.6$ & $2.0\pm3.7$ & $0.1\pm0.3$ & $\mathbf{0.0\pm0.0}$ \\

DiT + Speed
& $\mathbf{100\pm0}$ & $84\pm15$ & $73\pm14$ & $\mathbf{11\pm2}$ & $\mathbf{0.0\pm0.0}$ & $\mathbf{0.0\pm0.0}$ & $0.7\pm0.7$ & $0.4\pm1.0$ & $0.1\pm0.3$ & $\mathbf{0.0\pm0.0}$ \\

$\pi_{0.5}$
& $\mathbf{100\pm0}$ & $71\pm4$ & $58\pm3$ & $13\pm2$ & $\mathbf{0.0\pm0.0}$ & $0.1\pm0.3$ & $\mathbf{0.4\pm0.5}$ & $\mathbf{0.0\pm0.0}$ & $\mathbf{0.0\pm0.0}$ & $\mathbf{0.0\pm0.0}$ \\

$\pi_{0.5}$ + Speed
& $\mathbf{100\pm0}$ & $\mathbf{63\pm3}$ & $\mathbf{51\pm3}$ & $\mathbf{11\pm1}$ & $\mathbf{0.0\pm0.0}$ & $\mathbf{0.0\pm0.0}$ & $0.8\pm1.1$ & $\mathbf{0.0\pm0.0}$ & $\mathbf{0.0\pm0.0}$ & $\mathbf{0.0\pm0.0}$ \\
\bottomrule
\end{tabular}%
}
\vspace{-6pt}
\end{table}

\begin{table}[h]
\centering
\caption{Mini-Chair --- all individual metrics. Values are mean $\pm$ SD. Bold = best per column.}
\label{tab:policy_performance_handover_full}
\scriptsize
\setlength{\tabcolsep}{2pt}
\renewcommand{\arraystretch}{1.08}
\resizebox{\columnwidth}{!}{%
\begin{tabular}{@{}c@{}}
\begin{tabular}{@{}lccccccccc@{}}
\toprule
Policy
& \shortstack{Progress $\uparrow$\\(\%)} & \shortstack{Total time $\downarrow$\\(s)} & \shortstack{Robot time $\downarrow$\\(s)} & \shortstack{Human time $\downarrow$\\(s)} & \shortstack{Wrong\\commit. $\downarrow$} & \shortstack{S. Driver\\mistake $\downarrow$} & \shortstack{Screw\\mistake $\downarrow$} & \shortstack{Grasp\\mistake $\downarrow$} & \shortstack{HO\\mistake $\downarrow$} \\
\midrule
DiT
& $50\pm53$ & $318\pm56$ & $299\pm68$ & $18\pm15$ & $5.9\pm4.1$ & $\mathbf{0.0\pm0.0}$ & $\mathbf{0.0\pm0.0}$ & $\mathbf{0.0\pm0.0}$ & $0.3\pm0.7$ \\

DiT + Speed
& $50\pm53$ & $275\pm101$ & $264\pm107$ & $\mathbf{11\pm9}$ & $7.2\pm5.9$ & $\mathbf{0.0\pm0.0}$ & $0.1\pm0.3$ & $\mathbf{0.0\pm0.0}$ & $0.3\pm0.7$ \\

$\pi_{0.5}$
& $\mathbf{100\pm0}$ & $\mathbf{121\pm6}$ & $\mathbf{101\pm5}$ & $20\pm6$ & $0.1\pm0.3$ & $\mathbf{0.0\pm0.0}$ & $\mathbf{0.0\pm0.0}$ & $\mathbf{0.0\pm0.0}$ & $\mathbf{0.0\pm0.0}$ \\

$\pi_{0.5}$ + Speed
& $\mathbf{100\pm0}$ & $125\pm9$ & $102\pm9$ & $23\pm5$ & $\mathbf{0.0\pm0.0}$ & $\mathbf{0.0\pm0.0}$ & $\mathbf{0.0\pm0.0}$ & $\mathbf{0.0\pm0.0}$ & $0.1\pm0.3$ \\
\midrule
\end{tabular} \\[6pt]
\begin{tabular}{@{}lcccccc@{}}
\midrule
Policy
& \shortstack{S. Driver\\manip fail $\downarrow$} & \shortstack{Screw\\manip fail $\downarrow$} & \shortstack{Grasp\\manip fail $\downarrow$} & \shortstack{Failed\\HOs $\downarrow$} & \shortstack{HO\\collisions $\downarrow$} & \shortstack{False\\starts $\downarrow$} \\
\midrule
DiT
& $\mathbf{0.0\pm0.0}$ & $\mathbf{0.0\pm0.0}$ & $0.8\pm1.0$ & $0.1\pm0.3$ & $\mathbf{0.0\pm0.0}$ & $\mathbf{0.0\pm0.0}$ \\

DiT + Speed
& $0.3\pm0.5$ & $\mathbf{0.0\pm0.0}$ & $\mathbf{0.0\pm0.0}$ & $\mathbf{0.0\pm0.0}$ & $\mathbf{0.0\pm0.0}$ & $\mathbf{0.0\pm0.0}$ \\

$\pi_{0.5}$
& $\mathbf{0.0\pm0.0}$ & $\mathbf{0.0\pm0.0}$ & $\mathbf{0.0\pm0.0}$ & $\mathbf{0.0\pm0.0}$ & $\mathbf{0.0\pm0.0}$ & $0.1\pm0.3$ \\

$\pi_{0.5}$ + Speed
& $\mathbf{0.0\pm0.0}$ & $\mathbf{0.0\pm0.0}$ & $\mathbf{0.0\pm0.0}$ & $\mathbf{0.0\pm0.0}$ & $\mathbf{0.0\pm0.0}$ & $\mathbf{0.0\pm0.0}$ \\
\bottomrule
\end{tabular}
\end{tabular}%
}
\vspace{-6pt}
\end{table}

\subsection{False Commitment Analysis}
For each execution horizon $T_a$, we ran the sped-up $\pi_{0.5}$ at a handover boundary for five minutes eight times with an idle collaborator not in the training set. We define a false commitment as a full premature handover of the object the robot is holding and waiting drift as the cumulative Euclidean end-effector path length during the idle waiting trial. False commitments are a subset of wrong commitments, which we report in other experiments. We find that as $T_a$ increases, waiting drift and false commitments generally increase, while the condition with basin steering and pulling significantly reduces both issues as seen in Table \ref{tab:falsestarts}. We note that the small number of false commitments observed at $T_a = 4$ primarily arose after accumulated drift moved the robot outside the basin radius, eventually inducing a false commitment. Because FK-steering only works within $R_j$ of the nearest basin, if the robot slowly drifts away the policy will perform the same as the condition without steering. Hence, we attribute these false commitments to stochastic rollout variation after basin exit. We observe that for larger execution horizons $T_a$, the policy commands chunks with mixed idle and handover actions characteristic of action leakage.

\begin{table}[h]
\centering
\caption{False-commitment study: mean (SD) per condition and execution horizon.}
\label{tab:falsestarts}
\begin{tabular}{lrrrr}
\toprule
 & \multicolumn{4}{c}{Execution Horizon} \\
\cmidrule(lr){2-5}
Condition & $T_a{=}4$ & $T_a{=}8$ & $T_a{=}12$ & $T_a{=}16$ \\
\midrule
\multicolumn{5}{l}{\textbf{Waiting drift (m)}} \\
\textit{speed} & 1.04 (0.31) & 1.29 (0.24) & 2.20 (0.95) & 2.06 (1.56) \\
\textit{speed + pull} & 0.46 (0.40) & 0.40 (0.41) & 3.39 (2.51) & 3.19 (3.14) \\
\textit{speed + steer} & 1.48 (0.72) & 1.21 (0.37) & 1.76 (0.62) & 1.84 (0.76) \\
\textit{speed + steer + pull} & 0.49 (0.77) & 0.23 (0.44) & 0.24 (0.34) & 0.07 (0.02) \\
\midrule
\multicolumn{5}{l}{\textbf{False commitments}} \\
\textit{speed} & 0.1 (0.4) & 0.0 (0.0) & 0.2 (0.5) & 2.8 (3.3) \\
\textit{speed + pull} & 0.0 (0.0) & 0.0 (0.0) & 0.0 (0.0) & 1.5 (2.1) \\
\textit{speed + steer} & 0.4 (0.5) & 0.0 (0.0) & 0.2 (0.5) & 0.4 (0.5) \\
\textit{speed + steer + pull} & 0.0 (0.0) & 0.0 (0.0) & 0.1 (0.4) & 0.0 (0.0) \\
\bottomrule
\end{tabular}
\end{table}

\subsection{Jerk Analysis}
\begin{figure}[h]
    \begin{minipage}[b]{0.48\textwidth}
        \centering
        \includegraphics[width=\textwidth]{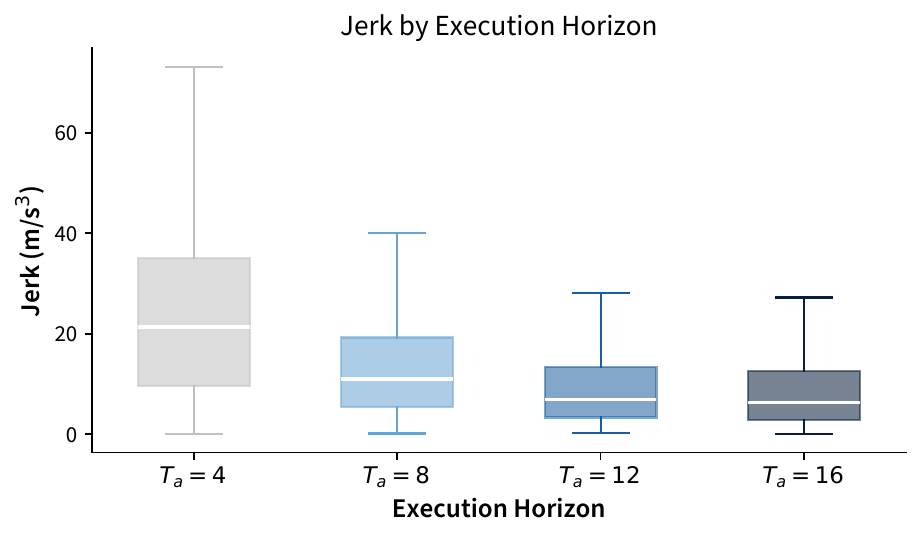}
        \caption{
            Per-timestep jerk measured at handover boundaries for each execution horizon.
        }
        \label{fig:jerk-box}
    \end{minipage}
    \hfill 
    \begin{minipage}[b]{0.48\textwidth}
        \centering
        \includegraphics[width=\textwidth]{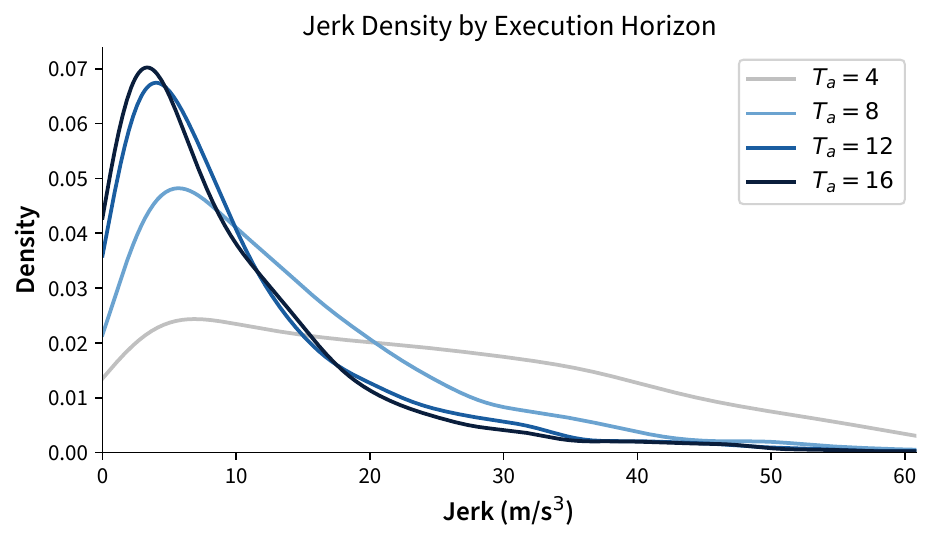}
        \caption{
            Gaussian KDE of per-timestep jerk magnitude (m/s$^3$).
        }
        \label{fig:jerk-density}
    \end{minipage}
\end{figure}

Since motion jerkiness was very noticeable to participants in our user study, we ran an additional experiment to quantify jerk across execution horizon choices. Ten handovers with a collaborator not present in the training set were recorded per execution horizon. We find that larger $T_a$ yields noticeably lower jerk as seen in Figure \ref{fig:jerk-box} and Figure \ref{fig:jerk-density}. Jerk from the end-effector position was estimated using finite-difference derivatives with respect to recorded timestamps. Specifically, for each rollout, we take the end-effector position $x_i \in \mathbb{R}^3$ at timestamp $t_i$ and apply NumPy's timestep-aware finite-difference operator iteratively and report $\| \hat{j_i} \|_2$ where $\hat{j}_i \approx d^3x/dt^3$. Interior derivatives are computed using centered finite differences and boundary samples are excluded. 

\subsection{User Evaluation}
Here we provide additional detail around the experimental procedure as well as extended study results across objective and subjective measures.

\subsubsection{Experimental Procedure}
Participants were first briefed on the high-level goals and structure of the experiment, followed by an informed consent process. After consenting, participants read a brief instruction manual for the Radio assembly task and then practiced the collaboration with the robot, where the robot was teleoperated by the experimenter. After training, participants completed the two experimental conditions, in a counterbalanced order. After each condition, participants completed the TLX, SUS, and fluency questionnaires. All experimental data, including robot camera images, were recorded for manual labeling of failures and task decomposition, which was performed by one of the authors. After completing both conditions, the participants filled out a brief demographics questionnaire, completed a short semi-structured interview regarding their experience with each condition, and finally were debriefed on the goals of the experiment. Participants were paid \$15 per hour for their participation and all participants completed the experiment in less than one hour.

\subsubsection{Extended User Study Results}
\begin{figure}[h]
    \centering
    \includegraphics[width=\textwidth]{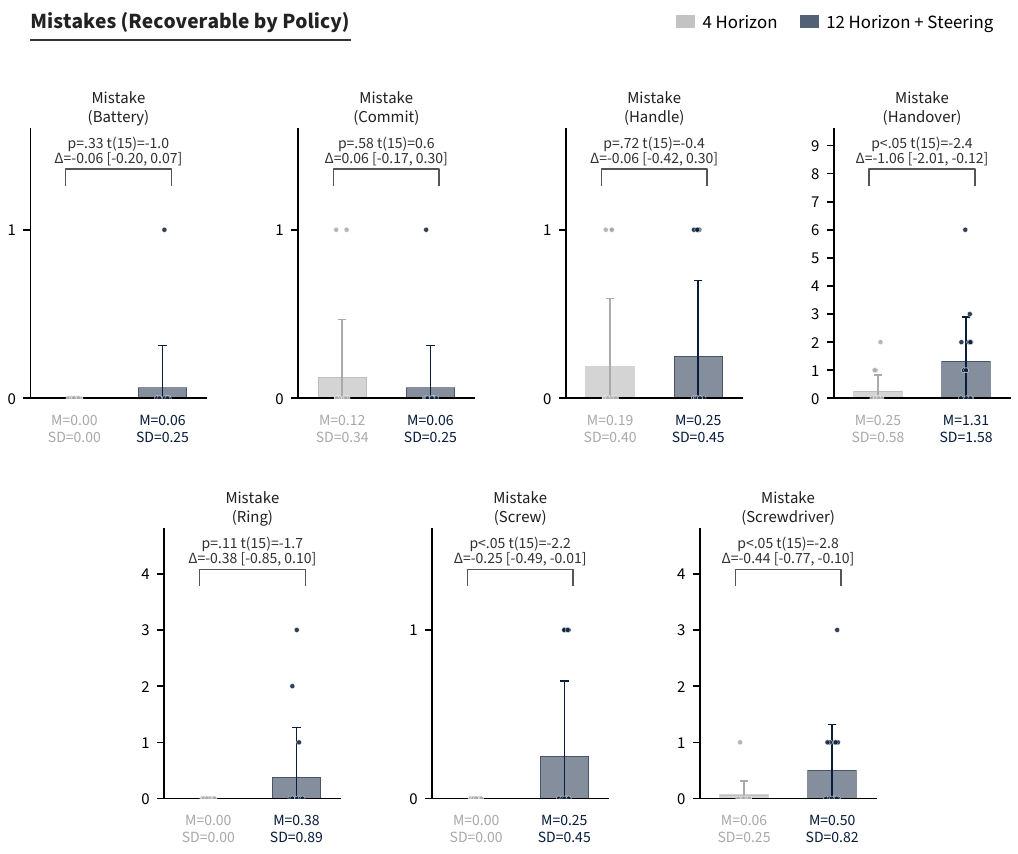}
    \caption{
        Robot policy mistakes that were recovered by the policy (only introducing inefficiency) broken down by the specific object and interaction. Ring refers to the speaker mesh cover the robot must place into the radio.
    }
    \label{fig:studymistakesonly}
\end{figure}

\begin{figure}[h]
    \centering
    \includegraphics[width=0.8\textwidth]{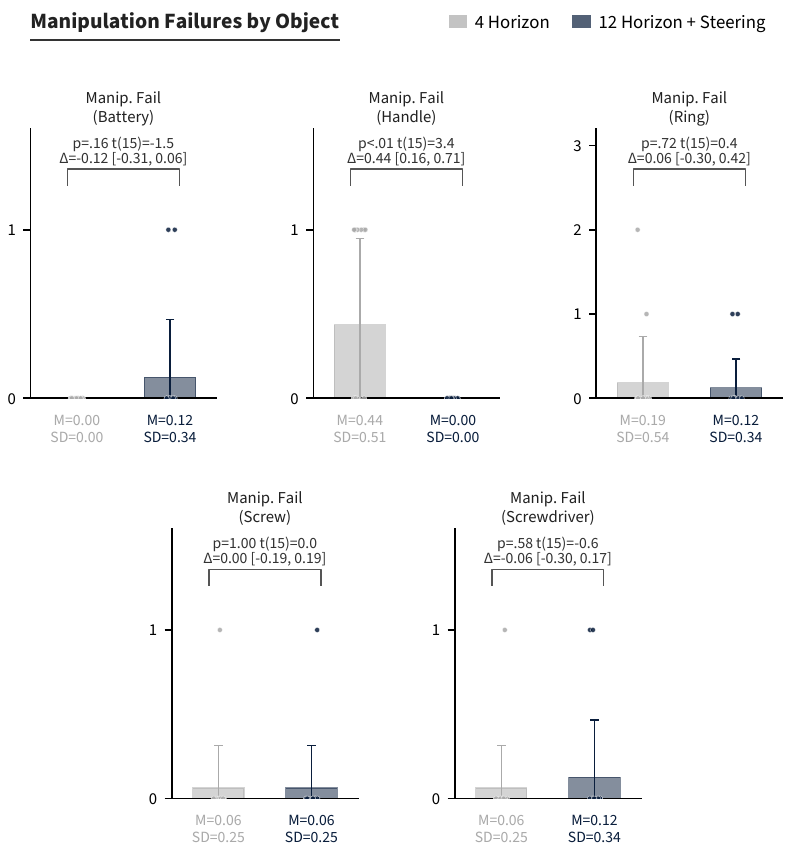}
    \caption{
        Manipulation failures broken down by the specific object and interaction. Failures require the robot to be physically reset. Ring refers to the speaker mesh cover the robot must place into the radio.
    }
    \label{fig:studyfailuresonly}
\end{figure}

We disaggregate user study mistakes and failures for each condition ($T_a = 4$ with speedup, and $T_a = 12$ with speedup, FK-steering, and basin pull) in Figure \ref{fig:studymistakesonly} and Figure \ref{fig:studyfailuresonly}, respectively. For conciseness, we will refer to FK-steering and basin pull together as steering. We selected these user study conditions based on the false commitment experiments and preliminary Radio rollouts. In preliminary Radio testing, the unsteered longer-horizon policy ($T_a = 12$) produced false commitments near handover transition states, whereas the shorter-horizon policy ($T_a = 4$) did not exhibit this behavior because frequent replanning limited execution of problematic action chunks that crossed transition boundaries. Steering mitigated these false commitments at $T_a = 12$, while preserving the smoother and more timely behavior associated with longer-horizon execution. We therefore compared the two practically viable mitigations for false commitments in the user study: frequent replanning with a short execution horizon, and longer-horizon execution with steering. Consistent with this design choice, we observed no false commitments in either user-study condition. We omitted additional ablation conditions to keep each user-study session under an hour and reduce participant fatigue.

As above, we record handover collisions whenever the robot touches the collaborator during handover. In most cases, this registered as only a light touch to the collaborator. Qualitatively, we observed that most collision cases occurred when participants extended their hand near the boundary of the policy's camera views and at a higher position than the demonstrators in the training set. Thus, most of the handover collisions we observed may have been avoidable with clearer handover-pose guidance or broader training data covering variable hand locations. While the handover collisions in our study were minor, they suggest that the longer-horizon condition trades off sensitivity to human handover behavior for improved task efficiency.

\begin{figure}[h]
    \centering
    \includegraphics[width=0.8\textwidth]{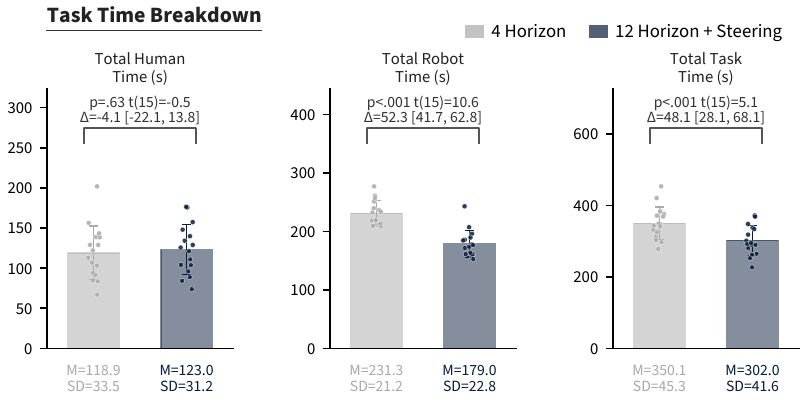}
    \caption{
        User study time results broken down by total time and times when the robot is moving.
    }
    \label{fig:studytimeonly}
\end{figure}

\begin{figure}[h]
    \centering
    \includegraphics[width=\textwidth]{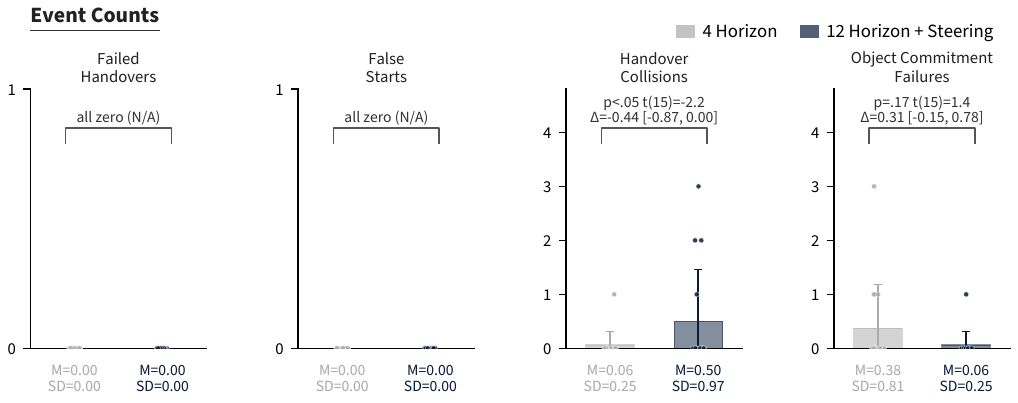}
    \caption{
        Counts of other collaboration related events in rollouts.
    }
    \label{fig:studycountsonly}
\end{figure}

\begin{figure}[h]
    \centering
    \includegraphics[width=0.8\textwidth]{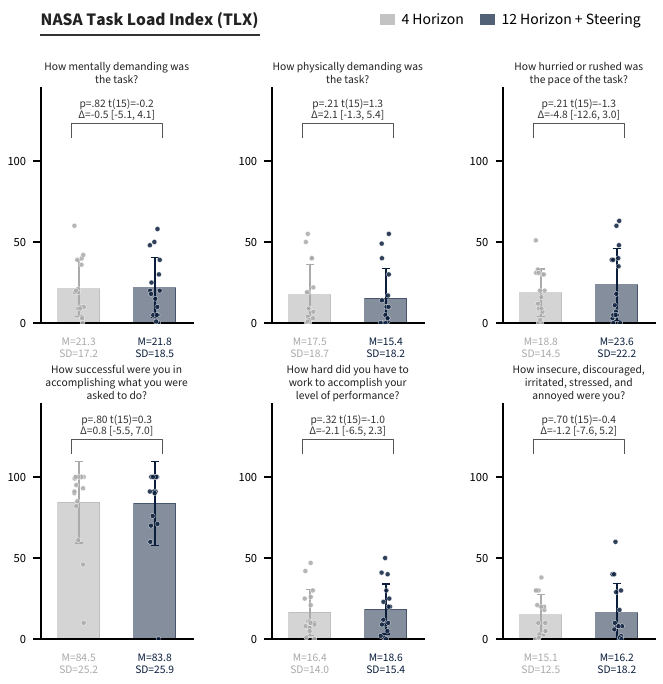}
    \caption{
       NASA-TLX broken down by subscale.
    }
    \label{fig:studytlxonly}
\end{figure}

\begin{figure}[h]
    \centering
    \includegraphics[width=\textwidth]{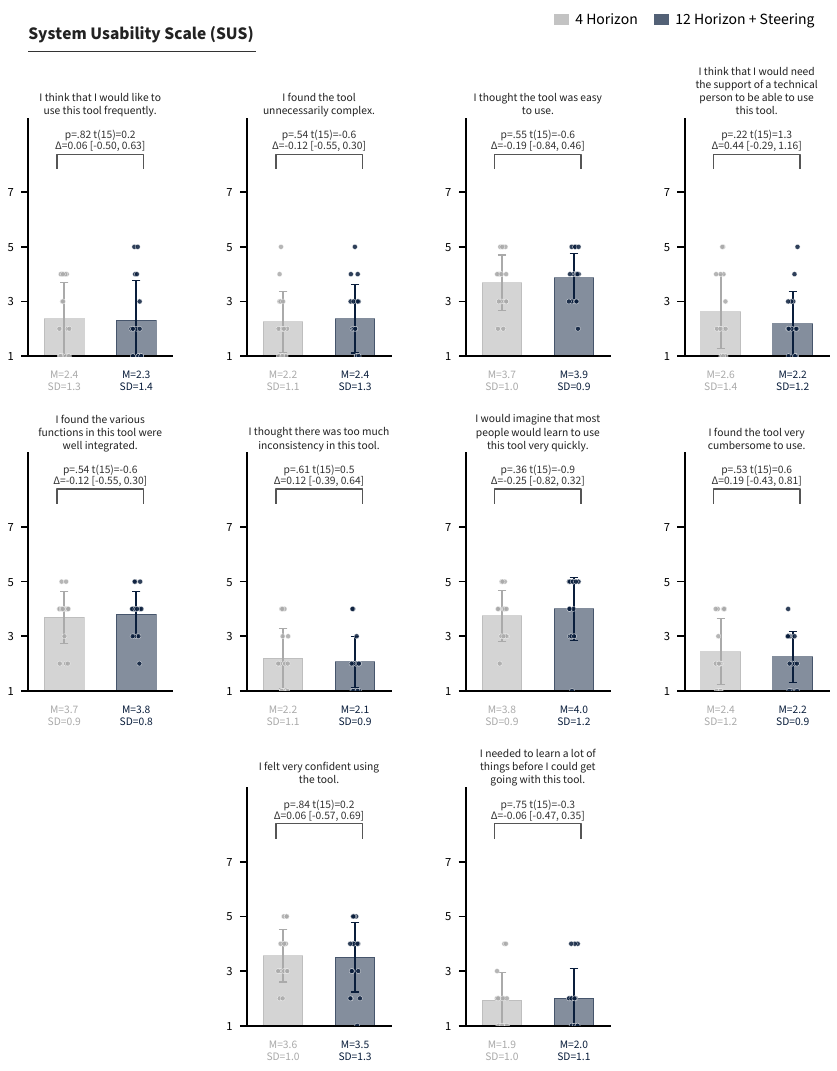}
    \caption{
       System Usability Scale (SUS) broken down by item.
    }
    \label{fig:studysusonly}
\end{figure}

\begin{figure}[h]
    \centering
    \includegraphics[width=\textwidth]{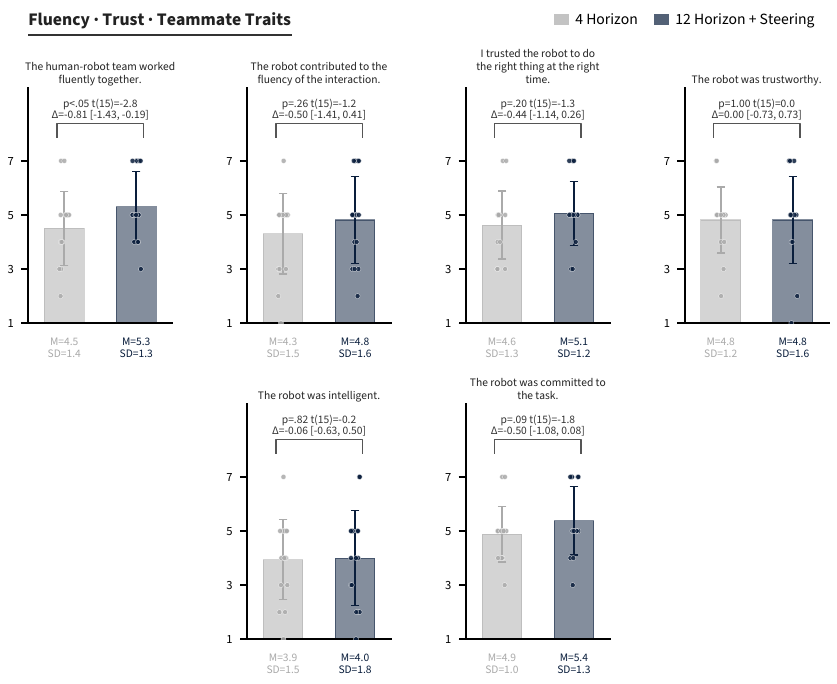}
    \caption{
       Human-Robot Fluency, Trust in Robot, and Positive Teammate Traits, broken down by question.
    }
    \label{fig:studyfluencyonly}
\end{figure}

\begin{figure}[h]
    \centering
    \includegraphics[width=\textwidth]{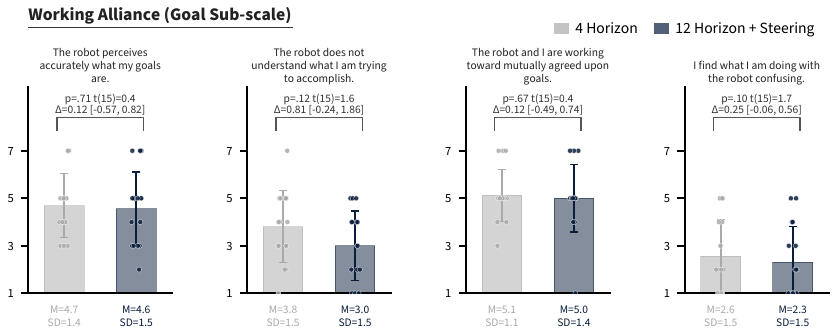}
    \caption{
       Working Alliance, broken down by question.
    }
    \label{fig:studyallianceonly}
\end{figure}

Figures \ref{fig:studytimeonly}-\ref{fig:studycountsonly} provide granular results of user study timing and events. These include task and robot time (Figure \ref{fig:studytimeonly}), mistakes (Figure \ref{fig:studymistakesonly}), failures (Figure \ref{fig:studyfailuresonly}), and other events (Fig. \ref{fig:studycountsonly}). In Figures \ref{fig:studytlxonly}, \ref{fig:studysusonly}, \ref{fig:studyfluencyonly}, and \ref{fig:studyallianceonly}, we break down the individual questions in the administered participant questionnaires. Each figure reports t-statistics, unadjusted p-values from paired t-tests, along with effect sizes and their 95\% confidence intervals.

For both conditions, our SUS scores were below the bottom 50 percent of ratings~\cite{bangor2008empirical}. We hypothesize this may be due to the failure rate of the VLAs (on the order of one failure per trial), which can degrade user experience and serves as inspiration for future work toward improving robustness of VLA HRC systems.

\end{document}